\documentclass{article}

\usepackage{microtype}
\usepackage{graphicx}
\usepackage{subcaption}
\usepackage{booktabs} 
\usepackage{multirow} 

\usepackage{hyperref}

\usepackage[preprint]{icml2026}

\usepackage{amsmath}
\usepackage{amssymb}
\usepackage{mathtools}
\usepackage{amsthm}

\usepackage{algorithm}
\usepackage{algorithmic}

\usepackage{tikz}

\usepackage[capitalize,noabbrev]{cleveref}

\theoremstyle{plain}

\theoremstyle{definition}

\theoremstyle{remark}

\usepackage[textsize=tiny]{todonotes}

\icmltitlerunning{Spiral RoPE: Rotate Your Rotary Positional Embeddings in the 2D Plane}

\begin{document}

\twocolumn[
  \icmltitle{Spiral RoPE \raisebox{-0.2em}{\includegraphics[height=1.1em]{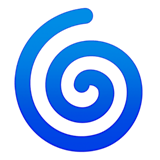}}: Rotate Your Rotary Positional Embeddings in the 2D Plane}



  \icmlsetsymbol{equal}{*}

  \begin{icmlauthorlist}
    \icmlauthor{Haoyu Liu}{ucb}
    \icmlauthor{Sucheng Ren}{jhu}
    \icmlauthor{Tingyu Zhu}{ucb}
    \icmlauthor{Peng Wang}{bytedance}
    \icmlauthor{Cihang Xie}{ucsc}
    \icmlauthor{Alan Yuille}{jhu}
    \icmlauthor{Zeyu Zheng}{ucb}
    \icmlauthor{Feng Wang}{jhu}
  \end{icmlauthorlist}

  \icmlaffiliation{ucb}{University of California, Berkeley}
  \icmlaffiliation{ucsc}{University of California, Santa Cruz}
  \icmlaffiliation{bytedance}{ByteDance}
  \icmlaffiliation{jhu}{John Hopkins University}

  \icmlcorrespondingauthor{Feng Wang}{wangf3014@gmail.com}

  \icmlkeywords{Vision Transformer, Positional Encoding, Rotary Position Embedding}

  \vskip 0.3in
]



\printAffiliationsAndNotice{%
  \textbf{Code and model checkpoints available at:}~%
  \url{https://github.com/huajianduzhuo-code/Spiral_RoPE}%
}

\begin{abstract}

Rotary Position Embedding (RoPE) is the de facto positional encoding in large language models due to its ability to encode relative positions and support length extrapolation. When adapted to vision transformers, the standard axial formulation decomposes two-dimensional (2D) spatial positions into horizontal and vertical components, implicitly restricting positional encoding to axis-aligned directions. We identify this directional constraint as a fundamental limitation of the standard axial 2D RoPE, which hinders the modeling of oblique spatial relationships that naturally exist in natural images. To lift this limitation, we propose \textbf{Spiral RoPE}, a simple yet effective extension that enables multi-directional positional encoding by partitioning embedding channels into multiple groups associated with uniformly distributed directions. Each group is rotated according to the projection of the patch position onto its corresponding direction, allowing spatial relationships to be encoded beyond the horizontal and vertical axes. Across a wide range of vision tasks including classification, segmentation, and generation, Spiral RoPE consistently improves performance. Qualitative analysis of attention maps further show that Spiral RoPE exhibits more concentrated activations on semantically relevant objects and better respects local object boundaries, highlighting the importance of multi-directional positional encoding in vision transformers.
\end{abstract}



\section{Introduction}
\label{sec:introduction}

\begin{figure}[t]
    \centering
    \includegraphics[width=\columnwidth]{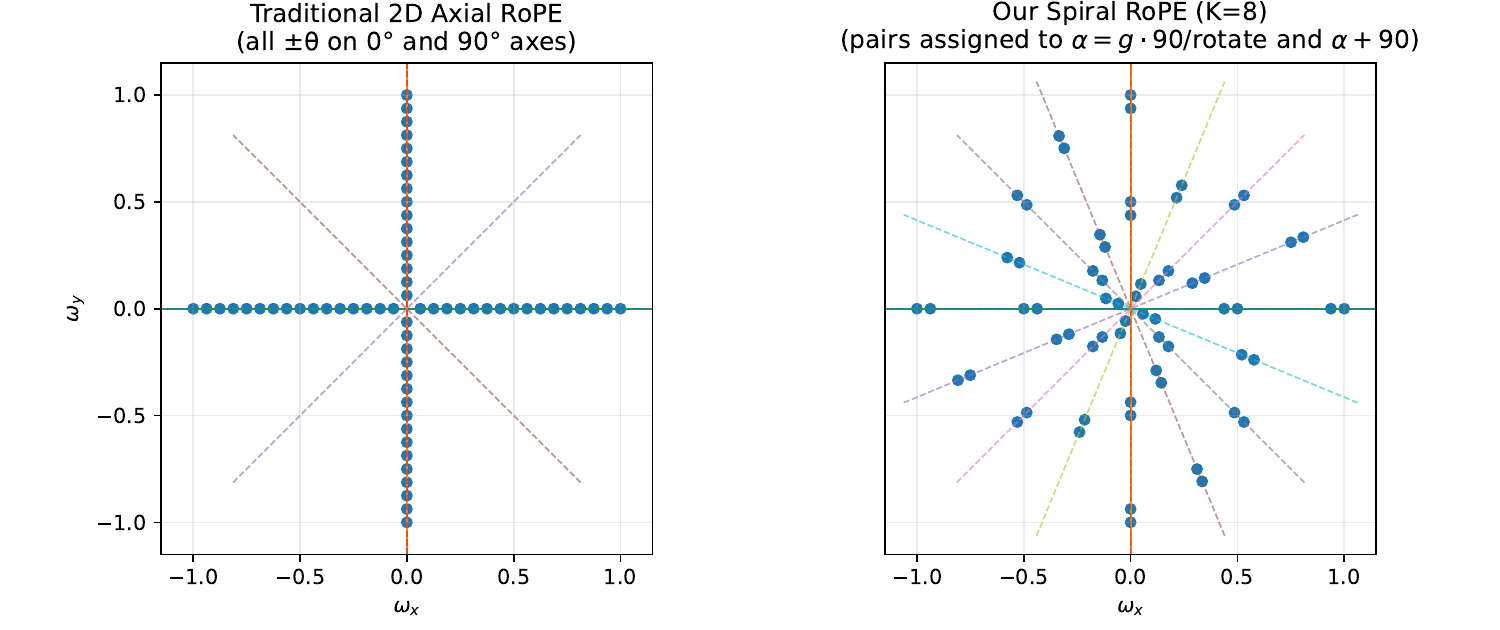}
    \caption{\textbf{\textit{Frequency support visualization}} comparing Axial 2D RoPE \textit{(left)} and Spiral RoPE \textit{(right)} with eight rotation groups ($K=8$). As shown, the Axial 2D RoPE places all frequencies on horizontal and vertical axes only, while our Spiral RoPE distributes frequencies across multiple directions in a spiral pattern, offering a broader directional coverage.}
    \label{fig:teaser}
    \vspace{-0.5cm}
\end{figure}

\begin{figure}[t]
    \centering
    \begin{subfigure}[b]{0.9\columnwidth}
        \centering
        \includegraphics[width=\linewidth]{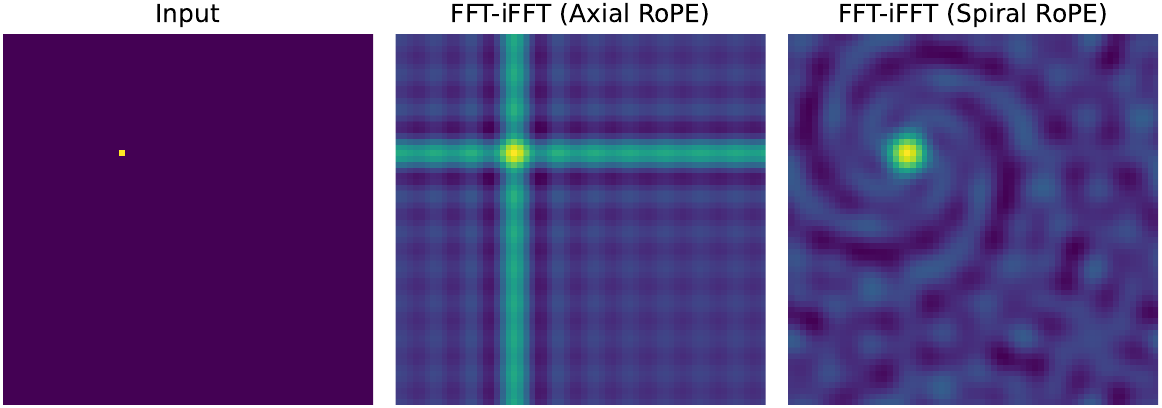}
        \caption{FFT-iFFT reconstruction of a single point}
    \end{subfigure}
    \hfill
    \begin{subfigure}[b]{0.9\columnwidth}
        \centering
        \includegraphics[width=\linewidth]{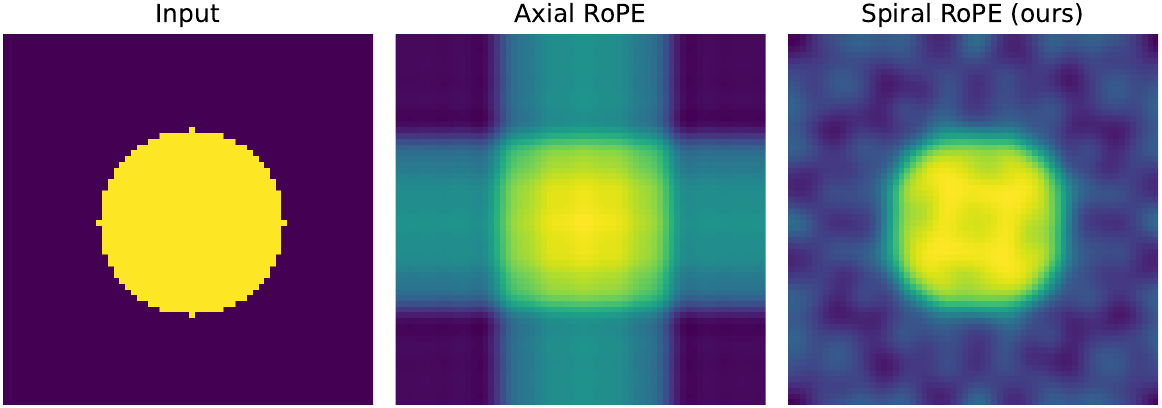}
        \caption{FFT-iFFT reconstruction of a circle}
    \end{subfigure}
    \caption{\textbf{\textit{Fourier reconstruction comparisons.}} Given binary input images \textit{(left)}, we retain only frequencies representable by each RoPE method and reconstruct via inverse-FFT. Axial 2D RoPE \textit{(middle)} produces artifacts along horizontal and vertical axes. Spiral RoPE achieves more faithful reconstruction due to broader directional coverage.}
    \vspace{-0.5cm}
    \label{fig:fourier}
\end{figure}

\begin{figure}[t]
    \centering
    \begin{subfigure}[b]{0.9\columnwidth}
        \centering
        \includegraphics[width=\linewidth]{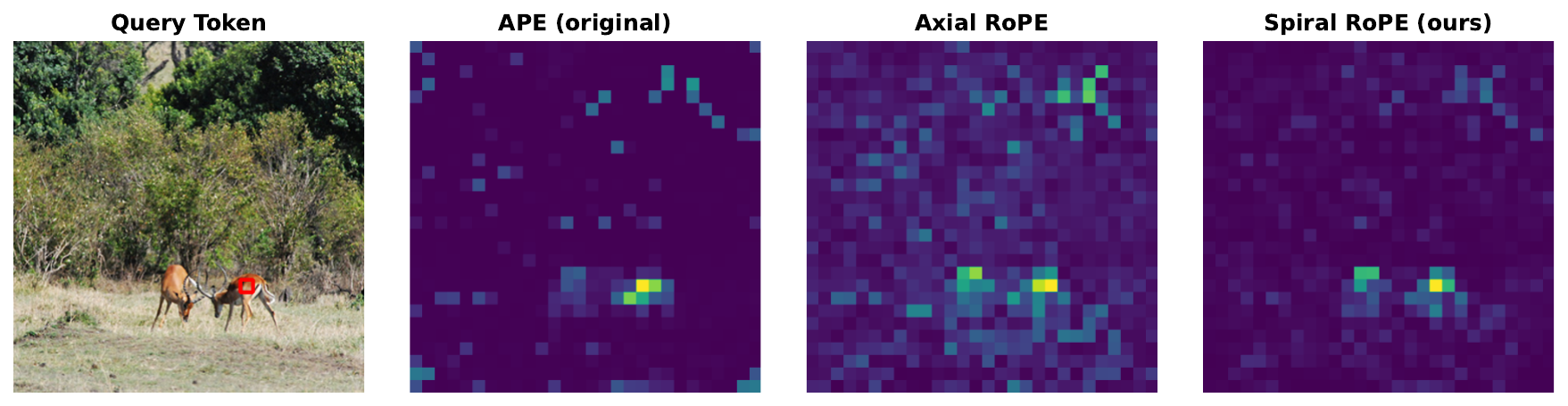}
    \end{subfigure}
    \\[0.6em]

    \begin{subfigure}[b]{0.9\columnwidth}
        \centering
        \includegraphics[width=\linewidth]{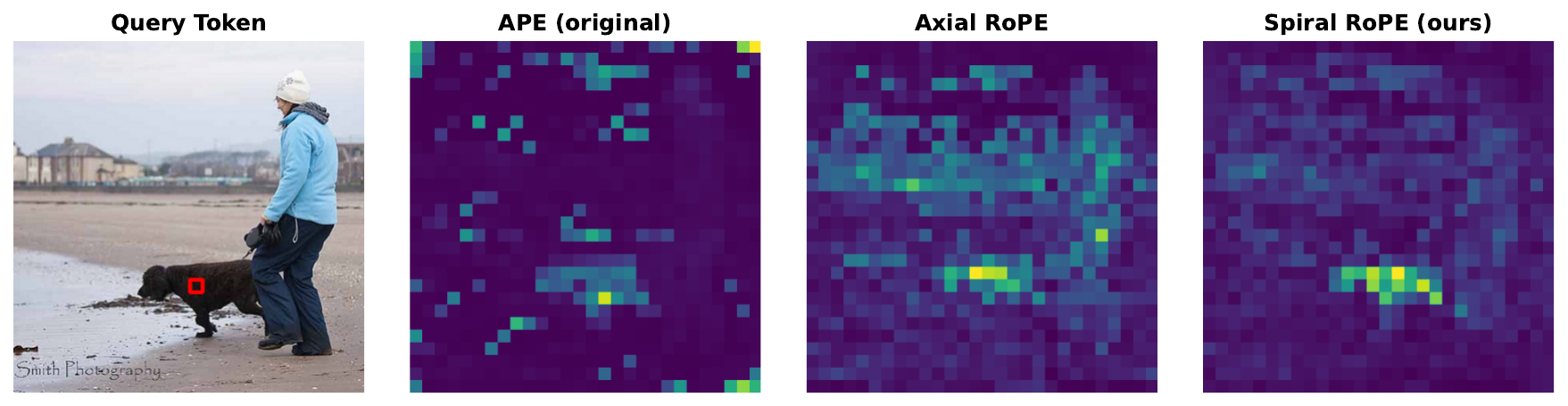}
    \end{subfigure}
    \\[0.6em]

    \begin{subfigure}[b]{0.9\columnwidth}
        \centering
        \includegraphics[width=\linewidth]{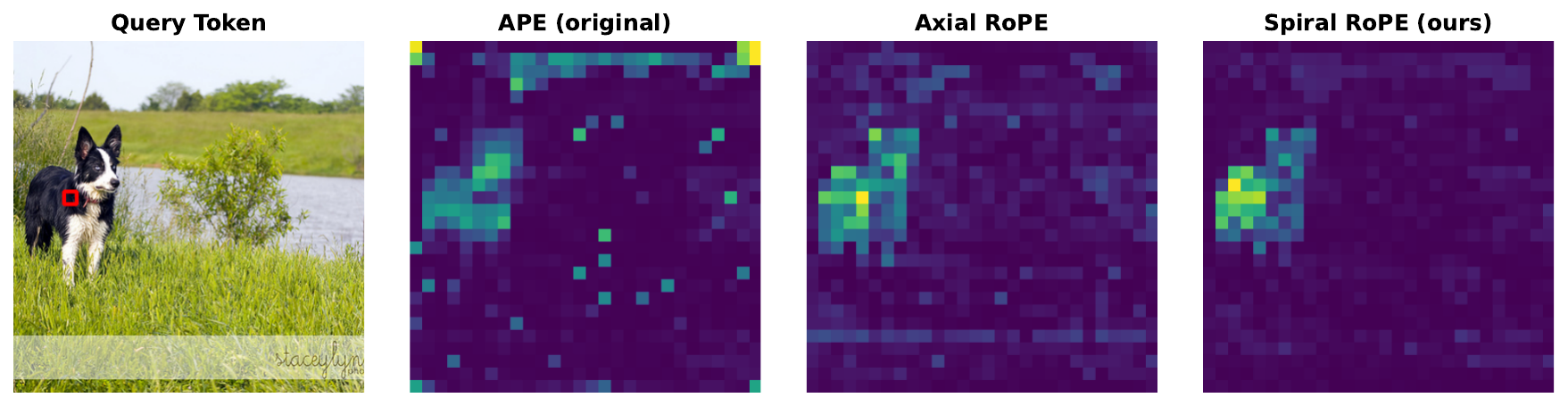}
    \end{subfigure}
    \caption{\textbf{\textit{Query-token attention visualization.}} The query token (marked by a red box) is located on a foreground object. Across diverse scenes, Spiral RoPE consistently produces sharper and more localized attention around the queried region compared to APE and Axial RoPE, indicating improved spatial alignment.}
    \label{fig:query_attn}
\end{figure}

Rotary Position Embedding (RoPE)~\cite{suRoFormerEnhancedTransformer2023} has become a cornerstone of modern language transformers, powering state-of-the-art large language models such as the LLaMA series~\cite{touvronLLaMAOpenEfficient2023, touvronLlama2Open2023, grattafioriLlama3Herd2024} and Qwen series~\cite{baiQwenTechnicalReport2023, yangQwen2TechnicalReport2024, yangQwen25TechnicalReport2025, yangQwen3TechnicalReport2025}. By encoding positional information through rotation operations on query and key vectors, RoPE naturally captures relative positions and enables effective length extrapolation. Such properties have been proven essential for scaling transformers to long contexts. In contrast, standard Vision Transformers (ViTs)~\cite{dosovitskiyImageWorth16x162021} predominantly rely on absolute positional embeddings (APE), which encode fixed position vectors that are added to patch embeddings. While simple and effective, APE has well-documented limitations: it struggles to generalize to resolutions unseen during training and provides no explicit mechanism for encoding relative spatial relationships between image patches.

Recent work has begun exploring RoPE for vision tasks~\cite{fangEVA02VisualRepresentation2024,luFiTFlexibleVision2024,luUnifiedIO2Scaling2023}, demonstrating promising results on image classification, dense prediction, and generation tasks. The standard approach extends 1D RoPE to 2D images by decomposing spatial positions into horizontal and vertical components. We term this approach as \textit{Axial 2D RoPE}. Half of the embedding dimensions are rotated based on the x-coordinate, while the other half uses the y-coordinate. However, this axial decomposition inherits a fundamental limitation: it can only encode positional relationships along the coordinate axes. when we visualize the frequency support of Axial 2D RoPE in the 2D Fourier domain, all frequencies lie exclusively on the horizontal and vertical axes. This means Axial 2D RoPE is insensitive to positional changes along diagonal or other oblique directions, which is a limitation given that natural images contain rich spatial structures in all orientations. To demonstrate this limitation concretely, we perform a Fourier reconstruction experiment in~\cref{fig:fourier}: given a binary image, we retain only the frequency components representable by each RoPE variant and reconstruct via inverse FFT (Fast Fourier Transform). Axial 2D RoPE produces visible artifacts along the coordinate axes, failing to faithfully reconstruct the input image structures such as circles.

In this work, we introduce \textbf{Spiral RoPE}, a new 2D rotary position embedding that extends directional coverage beyond the coordinate axes. Instead of splitting embeddings into just two axis-aligned groups, Spiral RoPE partitions them into $K$ groups corresponding to $K$ uniformly distributed directions in the 2D space. Each embedding group is rotated based on the patch position projected onto its assigned direction, enabling the model to explicitly encode positional relationships along diagonal and other oblique orientations. We design a grouped interleaved frequency assignment strategy that distributes the same number of distinct frequencies as Axial 2D RoPE across all $K$ directions, ensuring no loss in multi-scale encoding capacity. As shown in~\cref{fig:teaser}, the resulting frequency pattern forms a \textbf{\textit{spiral}} in the 2D frequency plane, providing broader directional coverage.

Notably, Spiral RoPE is simple to implement and introduces no additional computational overhead compared to Axial 2D RoPE, as both methods apply the same number of rotation operations with the same frequency budget, and our method does not introduce any additional learnable parameters. However, this simple change yields surprisingly significant and consistent improvements across diverse vision tasks. For example, on the standard ImageNet-1k~\cite{dengImageNetLargescaleHierarchical2009} classification task, Spiral RoPE improves test accuracy by +0.7\% for ViT-Large and +1.0\% for ViT-Base \textit{\textbf{with minimal cost}}. For semantic segmentation on ADE20k~\cite{zhouSceneParsingADE20K2017a}, our Spiral RoPE consistently achieves +2.2\% mIoU improvement for a ViT-Large backbone and +1.2\% mIoU for a ViT-Base. On class-conditional image generation with Diffusion Transformers (DiT)~\cite{peeblesScalableDiffusionModels2023}, Spiral RoPE reduces FID by 3.9--5.8 points across various model sizes. We also observe that Spiral RoPE demonstrates superior extrapolation capabilities to unseen resolutions, especially when the evaluation resolution is higher than the training resolution.

Beyond quantitative improvements, qualitative analyses further suggest that Spiral RoPE leads to more effective spatial representations. As illustrated in \cref{fig:fourier}, Spiral RoPE enables more faithful Fourier-domain reconstruction than Axial 2D RoPE, yielding reconstructions that more closely resemble the original images while avoiding characteristic axis-aligned artifacts. Consistent trends are also observed in attention map visualizations in \cref{fig:query_attn}, where we visualize attention patterns from individual query tokens on the foregroound object. Spiral RoPE produces sharper and more spatially coherent attention concentrated around the query location and semantically related regions, in contrast to the diffuse patterns produced by APE and the axis-biased activations of Axial RoPE. Together, these observations indicate that the directional constraints inherent in Axial 2D RoPE can limit spatial modeling in two-dimensional settings. By relaxing this restriction through a simple geometric extension, Spiral RoPE provides a more flexible positional encoding that supports richer spatial interactions, leading to improved spatial understanding in vision transformers. We hope this work can inspire generalizable architectural improvements for Vision Transforms.


\section{Related Work}
\label{sec:related_work}

\paragraph{Positional Encoding in Transformers.}
Transformers \cite{vaswaniAttentionAllYou2023} lack inherent positional awareness due to the permutation-equivariant nature of self-attention. To address this, the original Transformer used sinusoidal absolute position embeddings 
(APE), which add fixed position-dependent signals to input tokens. Learnable APE \cite{devlinBERTPretrainingDeep2019} replaces fixed sinusoids 
with trainable vectors, offering greater flexibility. However, APE methods encode absolute positions, which limits their ability to generalize to 
sequence lengths unseen during training. Relative position encodings address this limitation by encoding the distance between tokens rather than 
their absolute positions. Shaw et al.~\cite{shawSelfAttentionRelativePosition2018} introduced learnable relative position biases added to 
attention logits. T5 \cite{raffelExploringLimitsTransfer2023} further simplified this approach with a small number of learnable bias buckets. 
ALiBi \cite{pressTrainShortTest2022} proposed a non-learned linear bias that scales with distance, enabling length extrapolation in language 
models. When extending Transformers from 1D sequences to 2D visual data, vision Transformer (ViT) \cite{dosovitskiyImageWorth16x162021} applies Transformers to images by dividing them into patches, using learnable APE. DeiT \cite{touvronTrainingDataefficientImage2021} enabled efficient ViT training on ImageNet-1K, while Swin Transformer \cite{liuSwinTransformerHierarchical2021} introduced relative position bias within local windows. These developments highlight the importance of positional encoding design in vision models for handling varying resolutions and capturing spatial relationships.

\paragraph{Rotary Position Embedding.}
Rotary Position Embedding (RoPE) \cite{suRoFormerEnhancedTransformer2023} introduces a fundamentally different approach to position encoding by applying rotation operations directly to query and key vectors. Unlike additive position encodings, RoPE multiplies the query and key representations by rotation matrices whose angles are determined by the position index and a set of base frequencies. This design ensures that the dot product between query and key depends only on their relative positions, inherently encoding relative positional information. RoPE has become the de facto position encoding for modern large language models including LLaMA \cite{touvronLLaMAOpenEfficient2023} and Qwen \cite{yangQwen3TechnicalReport2025}, demonstrating excellent length extrapolation capabilities. Various extensions have been proposed to improve RoPE's ability to handle long sequences, including position interpolation \cite{chenExtendingContextWindow2023} which rescales positional indices at inference time and YaRN \cite{pengYaRNEfficientContext2023} which refines this strategy through frequency-aware scaling to improve stability.

\paragraph{2D RoPE for Vision.}
Extending RoPE to 2D vision data requires careful design choices. The straightforward approach, which we term axial 2D RoPE, splits the embedding dimension into two halves and applies 1D RoPE independently along the x and y axes. The 2D RoPE method has been adopted in many vision tasks, such as SAM 2 \cite{raviSAM2Segment2024} for image segmentation and Qwen-Image \cite{wuQwenImageTechnicalReport2025} for image generation. M-RoPE (Multimodal RoPE) \cite{wangQwen2VLEnhancingVisionLanguage2024} decomposes image indices into height, width, and temporal components to align visual patches with 1D text sequences. However, these axial approaches encode only horizontal and vertical relationships, without explicit awareness of diagonal or other directional spatial patterns. Heo et al.~\cite{heoRotaryPositionEmbedding2024} proposed RoPE-Mixed, treats the frequencies for both spatial axes as learnable parameters, allowing the network to adaptively "mix" axial signals to capture diagonal information. In multimodal settings, several works generalize RoPE to jointly handle tokens from different modalities. For instance, Circle-RoPE \cite{wangCircleRoPEConelikeDecoupled2025} project image token indices onto a ring that is orthogonal to the linear axis of text token indices to eliminate spurious cross-modal biases.


\section{Method}
\label{sec:method}


\subsection{Preliminaries: Rotary Position Embedding}

\subsubsection{1D RoPE}

Rotary Position Embedding \cite{suRoFormerEnhancedTransformer2023} encodes positional information through rotation operations. Given a $d$-dimensional query or key vector $\mathbf{x} \in \mathbb{R}^d$ at position $m$, RoPE applies a block-diagonal rotation matrix: 
\begin{equation}
    \text{RoPE}(\mathbf{x}, m) = \mathbf{R}_m \mathbf{x}
\end{equation}
where $\mathbf{R}_m$ is a block-diagonal matrix composed of $d/2$ 2D rotation matrices:
\begin{equation}
    \mathbf{R}_m = \begin{pmatrix}
        \mathbf{R}_m^{(0)} & & \\
        & \ddots & \\
        & & \mathbf{R}_m^{(d/2-1)}
    \end{pmatrix}
\end{equation}
with each $2 \times 2$ block being:
\begin{equation}
    \mathbf{R}_m^{(t)} = \begin{pmatrix}
        \cos(m\theta_t) & -\sin(m\theta_t) \\
        \sin(m\theta_t) & \cos(m\theta_t)
    \end{pmatrix}
\end{equation}

The frequency $\theta_t$ is typically set as $\theta_t = \theta_{\text{base}}^{-t/(d/2)}$ for $t = 0, 1, \ldots, d/2-1$, where $\theta_{\text{base}}$ is the base frequency that is commonly set to 10,000. This geometric progression of frequencies allows RoPE to encode positional dependencies at multiple scales.

A key property of RoPE is that the dot product between rotated vectors depends only on their relative position:
\begin{equation}
    \langle \text{RoPE}(\mathbf{q}, m), \text{RoPE}(\mathbf{k}, n) \rangle = \mathbf{q}^\top \mathbf{R}_{m-n} \mathbf{k}
    \label{eq:rope_relative}
\end{equation}
This property enables RoPE to implicitly encode relative positional information when calculating the attention matrix in the self-attention mechanism.

\subsubsection{Axial 2D RoPE}

RoPE was originally designed for 1D sequences. For 2D images, the standard approach is to extend RoPE by decomposing the spatial position into x and y coordinates. Given a patch at position $(p_x, p_y)$, the $d$-dimensional embedding $\mathbf{x} \in \mathbb{R}^d$ is split into two halves:
\begin{align}
    \mathbf{x}^{\text{(x)}} &= (x_0, x_1, \ldots, x_{d/2-1}) \\
    \mathbf{x}^{\text{(y)}} &= (x_{d/2}, x_{d/2+1}, \ldots, x_{d-1})
\end{align}

The first half is rotated using the x-coordinate, and the second half using the y-coordinate:
\begin{equation}
    \text{RoPE-2D}(\mathbf{x}, p_x, p_y) = \begin{pmatrix}
        \mathbf{R}_{p_x} \mathbf{x}^{\text{(x)}} \\
        \mathbf{R}_{p_y} \mathbf{x}^{\text{(y)}}
    \end{pmatrix}
\end{equation}

The rotation matrices $\mathbf{R}_{p_x}$ and $\mathbf{R}_{p_y}$ are typically constructed using the same set of frequencies. Each sub-embedding
$\mathbf{x}^{\text{(x)}}$ and $\mathbf{x}^{\text{(y)}}$ has dimensionality $d/2$ and thus can only accommodate $d/4$ distinct rotary frequencies, \emph{\emph{i.e.}}, half of those available in the original 1D RoPE.

Despite the good simplicity of transferring 1D RoPE to 2D inputs, this axial design only encodes positional relationships along the axis-aligned directions, leading to an inherent limitation. The x-coordinate and y-coordinate are processed independently, without any explicit encoding of diagonal or other directional relationships. Consider patches at positions $\mathbf{p}_1 = (0, 0)$ and $\mathbf{p}_2 = (1, 1)$, which are diagonally adjacent. With conventional 2D RoPE, the relative position encoding in the x-dimensions reflects $\Delta x = 1$, and in the y-dimensions reflects $\Delta y = 1$, but there is no direct encoding of the diagonal relationship.

\subsection{Spiral RoPE}

To address this limitation, we propose to extend 2D RoPE by applying additional \textbf{\textit{spatial rotations}} on the 2D panel. For example, we can directly encode the diagonal displacement $\sqrt{2}$ along the 45-degree direction, which can provide richer positional information. And more generally, we can encode positions along multiple uniformly distributed directions, which provides the model with a more complete description of spatial relationships.

Formally, let $K$ be the number of directions we want to encode. We uniformly distribute $K$ directions in the angular range $[0, \pi)$:
\begin{equation}
    \phi_k = \frac{k \pi}{K}, \quad k = 0, 1, \ldots, K-1
\end{equation}

Note that we only need to cover $[0, \pi)$ rather than $[0, 2\pi)$ because directions $\phi$ and $\phi + \pi$ are parallel, and RoPE naturally handles both positive and negative positions through its rotation mechanism.

For each direction $\phi_k$, we define a unit direction vector:
\begin{equation}
    \mathbf{u}_k = (\cos\phi_k, \sin\phi_k)
\end{equation}

Given a patch at 2D position $\mathbf{p} = (p_x, p_y)$, we compute its projected position along direction $k$ as:
\begin{equation}
    t_k(\mathbf{p}) = \mathbf{p} \cdot \mathbf{u}_k = p_x \cos\phi_k + p_y \sin\phi_k
    \label{eq:projection}
\end{equation}

We partition the $d$-dimensional embedding into $K$ equal groups, each with $d/K$ dimensions. The $k$-th group, denoted $\mathbf{x}^{(k)}$, is rotated using the projected position $t_k$:
\begin{equation}
    \text{Spiral RoPE}(\mathbf{x}, \mathbf{p}) = \begin{pmatrix}
        \mathbf{R}_{t_0(\mathbf{p})}^{(0)} \mathbf{x}^{(0)} \\
        \mathbf{R}_{t_1(\mathbf{p})}^{(1)} \mathbf{x}^{(1)} \\
        \vdots \\
        \mathbf{R}_{t_{K-1}(\mathbf{p})}^{(K-1)} \mathbf{x}^{(K-1)}
    \end{pmatrix}
\end{equation}
where $\mathbf{R}_{t}^{(k)}$ denotes the rotation matrix for direction $k$ with its specific frequency assignment.

A critical design choice in Spiral RoPE is how to assign rotation frequencies $\theta_i$ to different directions. A naive approach would independently assign $d/(2K)$ frequencies to each direction. However, this limits the number of distinct frequencies as $K$ increases, reducing the model's ability to encode positions at multiple scales.

To maximize frequency diversity, we adopt an interleaved frequency assignment strategy. We first compute a pool of $d/4$ base frequencies using the same formula as axial 2D RoPE:
\begin{equation}
    \theta_t = \theta_{\text{base}}^{-t/(d/4)}, \quad t = 0, 1, \ldots, d/4-1
\end{equation}
This gives us the same number of distinct frequencies as the conventional 2D RoPE baseline. 

To distribute these frequencies across directions, we adopt a grouped interleaved assignment strategy. We first group adjacent frequencies into pairs: $(\theta_0, \theta_1), (\theta_2, \theta_3), \ldots, (\theta_{d/4-2}, \theta_{d/4-1})$. We then assign these pairs to direction groups in a round-robin fashion. Specifically, we pair directions that are 90° apart (\emph{i.e.}, perpendicular), since they encode orthogonal spatial relationships. For direction pair $(k, k + K/2)$ where $k < K/2$, we assign frequency pairs:
\begin{equation}
    \Theta^{(k)} = \Theta^{(k+K/2)} = \{\theta_{2k}, \theta_{2k+1}, \theta_{2k+K}, \theta_{2k+K+1}, \ldots\}
\end{equation}

For example, with $K=4$ directions (0$^\circ$, 45$^\circ$, 90$^\circ$, 135$^\circ$) and $d=32$, the frequency assignment is:
\begin{itemize}
    \item $\theta_0, \theta_1, \theta_4, \theta_5$ are assigned to directions 0$^\circ$ and 90$^\circ$
    \item $\theta_2, \theta_3, \theta_6, \theta_7$ are assigned to directions 45$^\circ$ and 135$^\circ$
\end{itemize}

\cref{fig:teaser} visualizes the frequency assignment by plotting each frequency $\pm\theta_t$ as a point in the 2D frequency plane along the direction it encodes. For traditional 2D axial RoPE, all frequencies lie on the horizontal ($0^\circ$) and vertical ($90^\circ$) axes, corresponding to frequency vectors $(\pm\theta_t, 0)$ and $(0, \pm\theta_t)$. In contrast, Spiral RoPE with $K=8$ distributes frequencies across $K/2=4$ pairs of perpendicular directions: $0^\circ/90^\circ$, $22.5^\circ/112.5^\circ$, $45^\circ/135^\circ$, and $67.5^\circ/157.5^\circ$, which forms a \textit{spiral} pattern. This visualization illustrates how Spiral RoPE expands directional coverage beyond the coordinate axes while maintaining the same total number of frequencies.

This grouped interleaved assignment ensures that: (1) we utilize all $d/4$ distinct frequencies, matching the capacity of axial 2D RoPE; (2) perpendicular direction pairs share the same frequency set, analogous to how axial 2D RoPE uses the same frequencies for both x and y axes; and (3) each direction receives a mix of low and high frequencies from adjacent pairs, enabling multi-scale positional encoding within each direction.

Spiral RoPE preserves the relative position encoding property of RoPE: for patches at positions $\mathbf{p}$ and $\mathbf{p}'$, since $t_k(\mathbf{p}) - t_k(\mathbf{p}') = t_k(\mathbf{p} - \mathbf{p}')$, the attention score contribution depends only on the relative position $\mathbf{p} - \mathbf{p}'$. Furthermore, by uniformly distributing directional encodings, Spiral RoPE provides more isotropic spatial awareness compared to axial 2D RoPE, which treats horizontal and vertical directions preferentially. Note that Spiral RoPE can be implemented efficiently by precomputing the rotation matrices for all positions and directions, with minimal computational overhead compared to axial 2D RoPE.

To further illustrate the representation power of Spiral RoPE, we perform a 2D Fourier analysis comparing axial 2D RoPE with Spiral RoPE. We consider $d=1024$, which yields $d/4=256$ distinct RoPE frequencies. Given a binary image, we first apply 2D FFT to obtain its frequency-domain representation. We then retain only the frequency components that can be represented by each RoPE variant: for axial 2D RoPE, we keep frequencies along the horizontal and vertical axes $(\pm\theta_t, 0)$ and $(0, \pm\theta_t)$; for Spiral RoPE, we keep frequencies along all $K=8$ directions as illustrated in \cref{fig:teaser}. All other frequency components are masked to zero. Finally, we apply 2D iFFT to reconstruct the image and compare the result with the original input. \cref{fig:fourier} shows the reconstruction results for two input images: a single point and a circle on the $64 \times 64$ grid. The axial 2D RoPE reconstruction exhibits visible artifacts along the horizontal and vertical axes, as it can only represent frequency components in these two directions. In contrast, Spiral RoPE produces reconstructions that are closer to the original images, demonstrating its ability to capture a broader range of spatial patterns through multi-directional frequency coverage.


\section{Experiments}
\label{sec:experiments}

We conduct comprehensive experiments to validate the effectiveness of Spiral RoPE across multiple vision tasks: image classification on ImageNet-1k, semantic segmentation on ADE20k, and class-conditional image generation on ImageNet. Our experiments demonstrate that Spiral RoPE consistently improves performance over baselines and prior 2D RoPE methods.

\subsection{Image Classification}
\label{sec:experiments:classification}

\paragraph{Setup.} We evaluate Spiral RoPE on ImageNet-1k \cite{dengImageNetLargescaleHierarchical2009} classification using DeiT \cite{touvronTrainingDataefficientImage2021} as the base architecture. We train DeiT-Small, DeiT-Base and DeiT-Large models with and without Spiral RoPE on three target resolutions: $224 \times 224$, $384 \times 384$, and $448 \times 448$. Following \cite{heoRotaryPositionEmbedding2024}, we combine Spiral RoPE with the original absolute positional embedding (APE), as this combination has been shown to be effective. We use $K=16$ directions for Spiral RoPE with all frequencies $\theta_t$ scaled by a factor of 1.5 (i.e., $\theta_t = 1.5\theta_{\text{base}}^{-t/(d/4)}$). Models are pre-trained at $224 \times 224$ resolution for 300 epochs (Small and Base) or 200 epochs (Large), then fine-tuned at the target resolution for 20 epochs. We compare against the standard DeiT baseline and RoPE-Mixed \cite{heoRotaryPositionEmbedding2024}, which uses learnable per-head frequencies, in contrast to our method that uses pre-specified directions and frequencies. For evaluation, we report Top-1 accuracy on the validation set using the final checkpoint with exponential moving average (EMA) weights and full crop (crop ratio = 1.0). Additional training details are provided in Appendix~\ref{app:details}.

\paragraph{Results.} \cref{tab:classification} presents the classification results across different resolutions. Spiral RoPE consistently outperforms both the APE baseline and RoPE-Mixed across all configurations. These results demonstrate that the multi-directional encoding in Spiral RoPE provides richer positional information that benefits image classification, with larger improvements observed at higher resolutions.

\begin{table}[t]
    \caption{\textbf{\textit{ImageNet-1k classification accuracy (\%).}} Models are pre-trained at $224 \times 224$ and fine-tuned at each target resolution. Spiral RoPE consistently outperforms baselines across all configurations.}
    \label{tab:classification}
    \begin{center}
        \begin{small}
            \begin{sc}
                \begin{tabular}{llccc}
                    \toprule
                    Model & Method & 224 & 384 & 448 \\
                    \midrule
                    \multirow{3}{*}{DeiT-S} 
                    & APE & 79.11 & 81.88 & 82.24 \\
                    & RoPE-Mixed & \textbf{80.48} & 82.64 & 82.99 \\
                    & Spiral RoPE (ours) & 80.39 & \textbf{83.04} & \textbf{83.15} \\
                    \midrule
                    \multirow{3}{*}{DeiT-B} 
                    & APE & 82.36 & 84.16 & 84.29 \\
                    & RoPE-Mixed & 83.15 & 84.73 & 84.84 \\
                    & Spiral RoPE (ours) & \textbf{83.39} & \textbf{85.04} & \textbf{85.19} \\
                    \midrule
                    \multirow{3}{*}{DeiT-L} 
                    & APE & 83.24 & 84.92 & 85.13 \\
                    & RoPE-Mixed & 83.51 & 85.14 & 85.40 \\
                    & Spiral RoPE (ours) & \textbf{83.97} & \textbf{85.48} & \textbf{85.67} \\
                    \bottomrule
                \end{tabular}
            \end{sc}
        \end{small}
    \end{center}
    \vskip -0.1in
\end{table}

\paragraph{Extrapolation Robustness.} We further evaluate robustness to input resolutions unseen during training. Using models trained and fine-tuned at $224 \times 224$ resolution, we evaluate performance across resolutions ranging from $144 \times 144$ to $512 \times 512$. Other evaluation settings are the same as before. For resolution extrapolation, APE employs bicubic interpolation to resize the learned position vectors from training to evaluation resolution. RoPE-based methods extend position indices to the new grid size while preserving the same frequency patterns, enabling smooth extrapolation through the inherent continuity of trigonometric functions.

\begin{figure}[t]
    \centering
    \begin{subfigure}[b]{0.9\columnwidth}
        \centering
        \includegraphics[width=\linewidth]{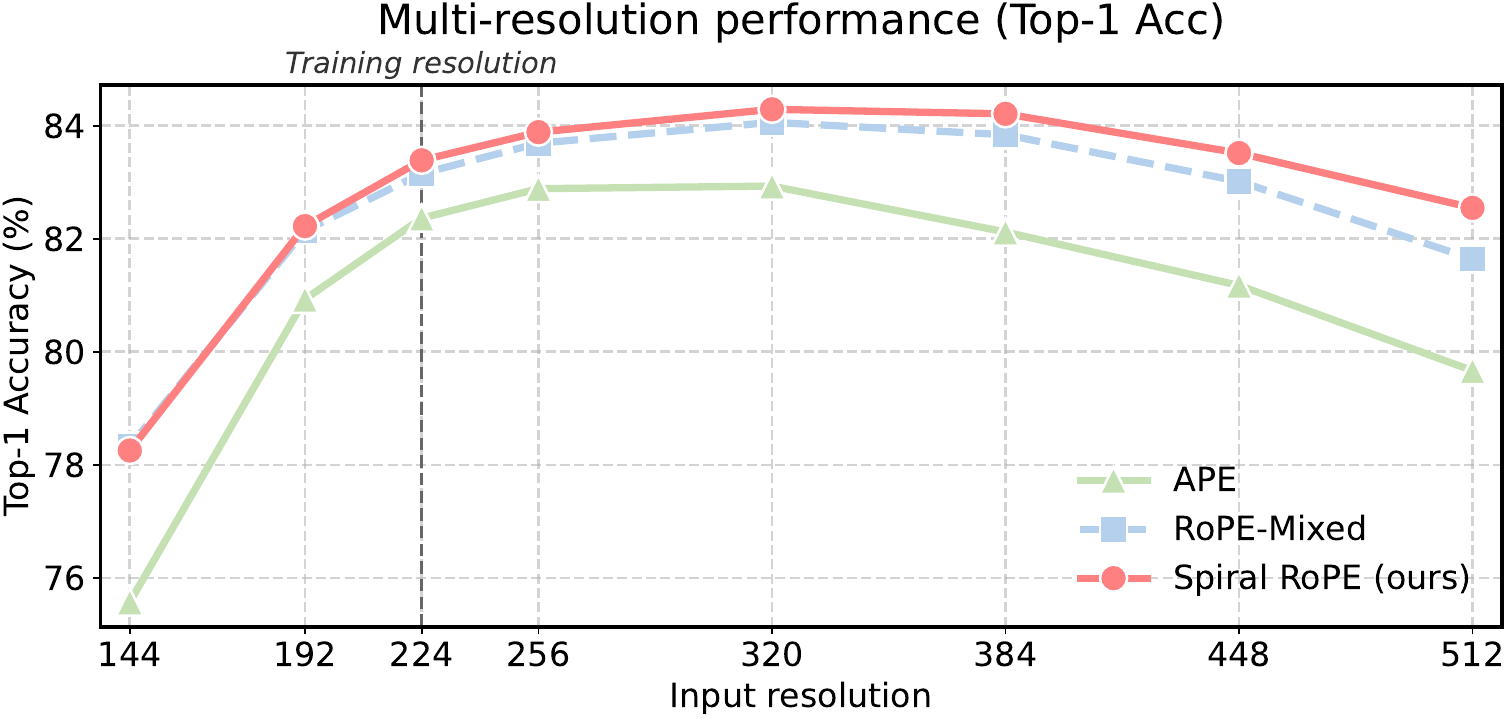}
        \caption{DeiT-Base}
    \end{subfigure}
    \\[0.3em]
    \begin{subfigure}[b]{0.9\columnwidth}
        \centering
        \includegraphics[width=\linewidth]{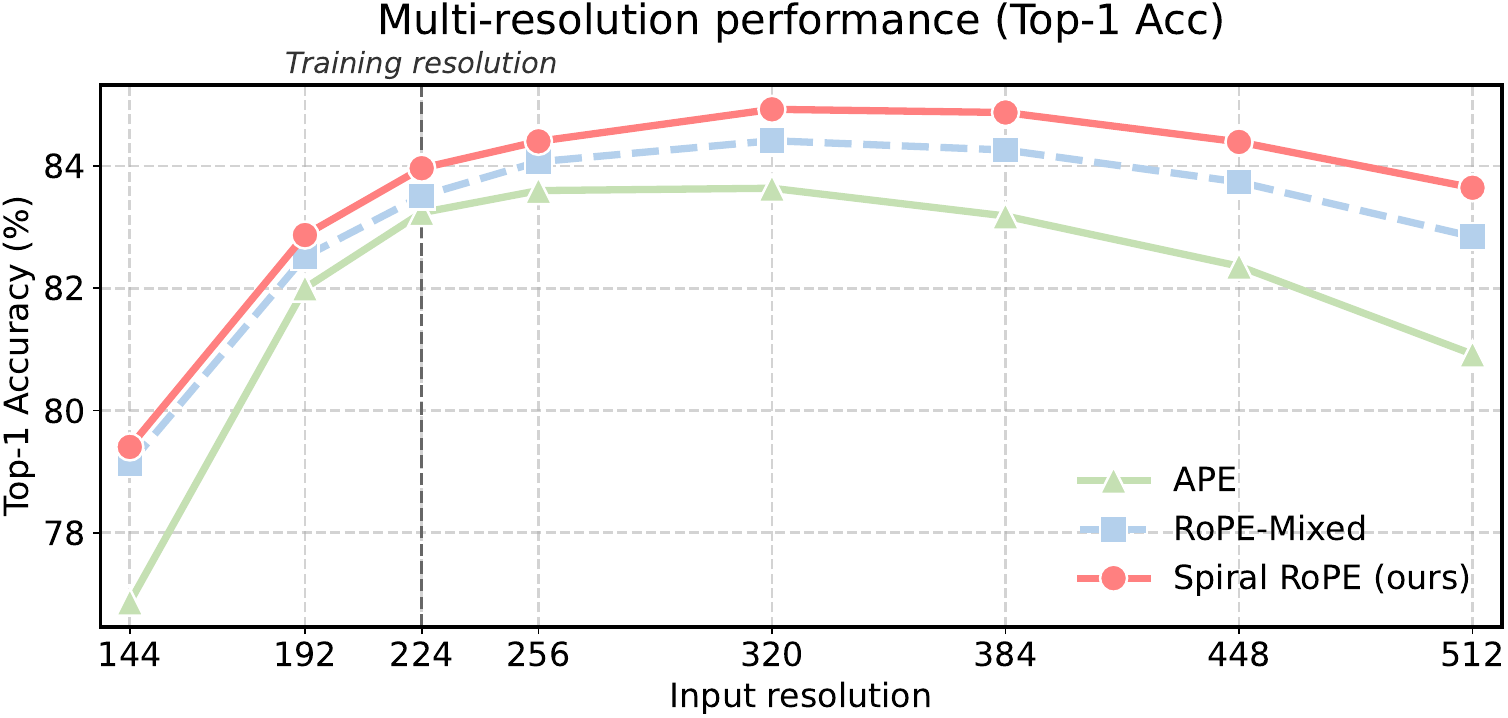}
        \caption{DeiT-Large}
    \end{subfigure}
    \caption{\textbf{\textit{Multi-resolution performance on ImageNet-1k}}. Models are trained at $224 \times 224$ and evaluated across resolutions from $144 \times 144$ to $512 \times 512$. Spiral RoPE demonstrates superior robustness to resolution changes compared to APE and RoPE-Mixed baselines across both model sizes.}
    \label{fig:multires}
\end{figure}

\cref{fig:multires} shows the extrapolation robustness results for both DeiT-Base and DeiT-Large models. Both Spiral RoPE and RoPE-Mixed consistently outperform APE across all resolutions and model sizes, demonstrating the advantage of rotary position embeddings for resolution generalization. At lower resolutions near the training resolution, Spiral RoPE and RoPE-Mixed perform comparably. However, at higher resolutions (\emph{e.g.}, $384 \times 384$ to $512 \times 512$), our Spiral RoPE exhibits a clear advantage over RoPE-Mixed across both model sizes, suggesting that our method with pre-specified directions and fixed sinusoidal frequencies achieves better extrapolation than RoPE-Mixed with learned frequencies when evaluating on unseen resolutions.

\subsection{Semantic Segmentation}
\label{sec:experiments:segmentation}

\paragraph{Setup.} We evaluate Spiral RoPE on semantic segmentation using the ADE20k dataset \cite{zhouSceneParsingADE20K2017a, zhouSemanticUnderstandingScenes2018a}, which contains 20,000 training images and 2,000 validation images across 150 semantic categories. We use UperNet \cite{xiaoUnifiedPerceptualParsing2018} as the segmentation framework with ViT training recipe, and we initialize the backbone with the ImageNet pre-trained weights at $224 \times 224$. The segmentation models are trained at $512 \times 512$ resolution for 160k iterations with a batch size of 16. For evaluation, we perform single-scale testing on the validation set and report mean Intersection over Union (mIoU), mean class accuracy (mAcc), and overall pixel accuracy (aAcc). Additional training details are provided in Appendix~\ref{app:details}.

\paragraph{Results.} \cref{tab:segmentation} summarizes semantic segmentation results on ADE20k, where Spiral RoPE consistently achieves significant improvements over the original APE baselines. Remarkably, we observe that the DeiT-Large model attains a notable mIoU of 49.12\%, substantially outperforming the original baseline of 46.91\% at virtually no additional computational cost. Moreover, the gains brought by Spiral RoPE on the ADE20K semantic segmentation task are intuitively more pronounced than those observed in classification (\emph{e.g.}, 3.06\% vs. 0.73\% accuracy improvement), suggesting that for dense prediction tasks such as semantic segmentation where reasoning is more complex and higher input resolutions are required, a well-designed relative position embedding can more effectively facilitate the modeling of spatial dependencies, demonstrating that Spiral RoPE consistently delivers larger benefits on more challenging tasks.

\begin{table}[t]
    \caption{\textbf{\textit{ADE20k semantic segmentation results.}} All models are pretrained on ImageNet-1k and then fintuned with an UperNet segmentation head for 160k iterations at $512 \times 512$ resolution.}
    \label{tab:segmentation}
    \begin{center}
        \begin{small}
            \begin{sc}
                \begin{tabular}{llccc}
                    \toprule
                    Backbone & Method & mIoU & mAcc & aAcc \\
                    \midrule
                    \multirow{2}{*}{DeiT-S} 
                    & APE & 43.72 & 54.33 & 81.07 \\
                    & Spiral RoPE & \textbf{45.44} & \textbf{56.33} & \textbf{81.61} \\
                    \midrule
                    \multirow{2}{*}{DeiT-B} 
                    & APE & 46.89 & 57.28 & 82.30 \\
                    & Spiral RoPE & \textbf{48.11} & \textbf{59.07} & \textbf{82.76} \\
                    \midrule
                    \multirow{2}{*}{DeiT-L} 
                    & APE & 46.91 & 57.05 & 82.36 \\
                    & Spiral RoPE & \textbf{49.12} & \textbf{60.11} & \textbf{83.40} \\
                    \bottomrule
                \end{tabular}
            \end{sc}
        \end{small}
    \end{center}
    \vskip -0.1in
\end{table}

\subsection{Image Generation}

\paragraph{Setup.} We also evaluate Spiral RoPE on class-conditional image generation using Diffusion Transformers (DiT) \cite{peeblesScalableDiffusionModels2023}. We train DiT models of various sizes (S, B, L, XL) with patch sizes of 2 (denoted as /2 in the original paper) on ImageNet $256 \times 256$. The /2 models process $128 \times 128$ latent patches ($16 \times 16$ tokens after patchification). We use $K=8$ directions for S/B/L models and $K=6$ for XL models. All models are trained for 400k steps with the original training recipe of DiT~\cite{peeblesScalableDiffusionModels2023}. For evaluation, we generate 50k samples using DDPM sampling with 250 steps and classifier-free guidance scale of 1.0, and compute Fr\'{e}chet Inception Distance (FID) \cite{heuselGANsTrainedTwo2018} against the ImageNet validation set statistics. Additional training details are provided in Appendix~\ref{app:details}.

\paragraph{Results.} \cref{tab:generation} presents the FID scores. The results suggest that Spiral RoPE is most effective when the sequence length is sufficiently large for the multi-directional encoding to capture meaningful spatial relationships.

\begin{table}[t]
    \caption{\textbf{\textit{Class-conditional image generation}} on ImageNet $256 \times 256$ using DiT. We compare models trained for 400k steps with the standard training hyperparameters of DiT. We generate 50k samples using DDPM sampling with 250 steps and classifier-free guidance scale of 1.0. Baseline results with supermarks * are taken from U-DiTs~\cite{tianUDiTsDownsampleTokens2024}.}
    \label{tab:generation}
    \begin{center}
        \resizebox{\columnwidth}{!}{
            \begin{sc}
                \begin{tabular}{ll@{\hskip 6pt}c@{\hskip 6pt}c@{\hskip 6pt}c@{\hskip 6pt}c@{\hskip 6pt}c}
                    \toprule
                    Model & Method & FID$\downarrow$ & sFID$\downarrow$ & IS$\uparrow$ & Prec.$\uparrow$ & Rec.$\uparrow$ \\
                    \midrule
                    \multirow{2}{*}{DiT-S/2} 
                    & APE$^*$ & 67.40 & 11.93 & 20.44 & 0.368 & 0.559 \\
                    & Spiral RoPE & \textbf{63.33} & \textbf{11.91} & \textbf{21.96} & \textbf{0.386} & \textbf{0.574} \\
                    \midrule
                    \multirow{2}{*}{DiT-B/2} 
                    & APE$^*$ & 42.84 & 8.24 & 33.66 & 0.491 & 0.629 \\
                    & Spiral RoPE & \textbf{37.74} & \textbf{8.23} & \textbf{38.31} & \textbf{0.520} & \textbf{0.630} \\
                    \midrule
                    \multirow{2}{*}{DiT-L/2} 
                    & APE$^*$ & 23.27 & \textbf{6.35} & 59.63 & 0.611 & \textbf{0.635} \\
                    & Spiral RoPE & \textbf{19.02} & 6.36 & \textbf{69.49} & \textbf{0.638} & 0.632 \\
                    \midrule
                    \multirow{2}{*}{DiT-XL/2} 
                    & APE$^*$ & 20.05 & 6.25 & 66.74 & 0.632 & 0.629 \\
                    & Spiral RoPE & \textbf{15.55} & \textbf{5.87} & \textbf{79.67} & \textbf{0.662} & \textbf{0.631} \\
                    \bottomrule
                \end{tabular}
            \end{sc}
        }
    \end{center}
    \vskip -0.1in
\end{table}

\begin{figure}[t]
    \centering
    \includegraphics[width=\columnwidth]{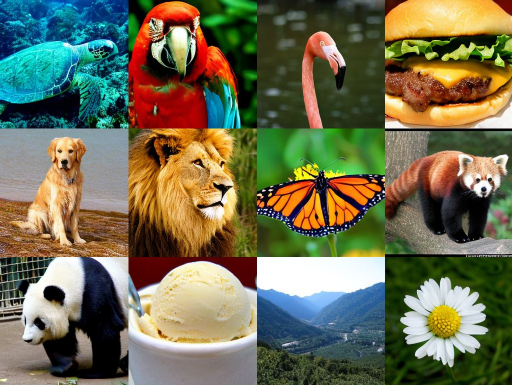}
    \caption{\textbf{\textit{Generated samples}} from our DiT-XL/2 model with Spiral RoPE trained for 7M steps on ImageNet $256 \times 256$. Images are generated using classifier-free guidance with scale 4.0.}
    \label{fig:generation_samples}
\end{figure}

\paragraph{Extended Training.} To further demonstrate the effectiveness of Spiral RoPE when scaled to longer training, we conduct experiments following the setup of SiT~\cite{maSiTExploringFlow2024}. Specifically, we train the XL/2 models for 7 million steps on ImageNet $256 \times 256$ and evaluate using classifier-free guidance with scale 1.5. \cref{tab:generation_extended} compares our Spiral RoPE against the original DiT baseline and SiT. Our Spiral RoPE achieves an FID of 1.74, outperforming both DiT (2.27) and SiT (2.06). \cref{fig:generation_samples} shows qualitative samples generated by our model.

\begin{table}[t]
    \caption{\textbf{\textit{Extended training comparison}} on ImageNet $256 \times 256$. The Spiral RoPE model uses DiT-XL/2 architecture trained for 7M steps. We use classifier-free guidance with scale 1.5. Baseline results are from SiT~\cite{maSiTExploringFlow2024}.}
    \label{tab:generation_extended}
    \begin{center}
        \resizebox{\columnwidth}{!}{
            \begin{sc}
                \begin{tabular}{l@{\hskip 8pt}c@{\hskip 8pt}c@{\hskip 8pt}c@{\hskip 8pt}c@{\hskip 8pt}c}
                    \toprule
                    Method & FID$\downarrow$ & sFID$\downarrow$ & IS$\uparrow$ & Prec.$\uparrow$ & Rec.$\uparrow$ \\
                    \midrule
                    DiT & 2.27 & 4.60 & 278.24 & 0.83 & 0.57 \\
                    SiT & 2.06 & 4.49 & 277.50 & 0.83 & 0.59 \\
                    Spiral RoPE (ours) & \textbf{1.74} & \textbf{4.49} & \textbf{279.64} & \textbf{0.83} & \textbf{0.60} \\
                    \bottomrule
                \end{tabular}
            \end{sc}
        }
    \end{center}
    \vskip -0.1in
\end{table}

\section{Analysis}
\label{sec:analysis}

In this section, we provide further analysis of Spiral RoPE, including visualization of attention maps and ablation studies on key hyperparameters.

\subsection{Visualization of Attention Maps}

To qualitatively understand the differences between positional encoding methods, we visualize attention maps from DeiT-Base models fine-tuned at $448 \times 448$ resolution. We compare three variants: the original APE baseline, APE with axial 2D RoPE, and APE with Spiral RoPE ($K=16$).

\begin{figure}[t]
    \centering

    \begin{subfigure}[b]{0.9\columnwidth}
        \centering
        \includegraphics[width=\linewidth]{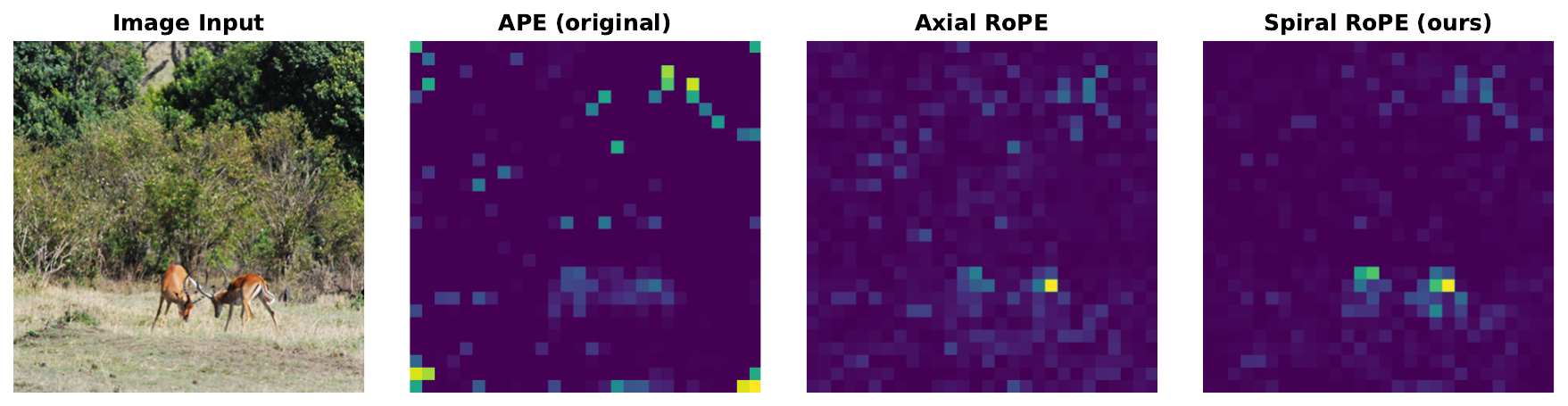}
    \end{subfigure}
    \\[0.6em]

    \begin{subfigure}[b]{0.9\columnwidth}
        \centering
        \includegraphics[width=\linewidth]{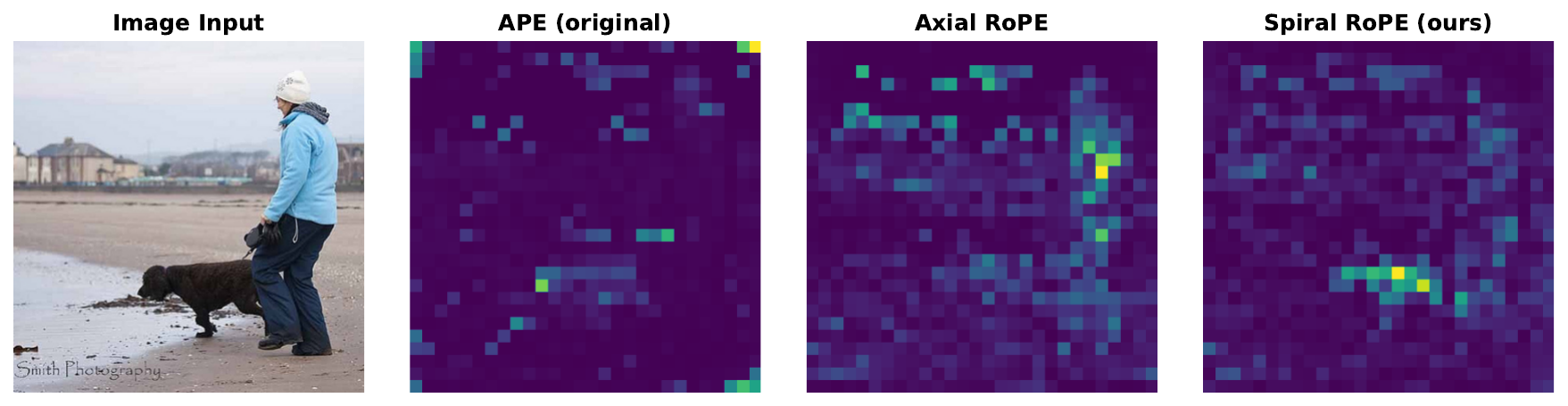}
    \end{subfigure}
    \\[0.6em]

    \begin{subfigure}[b]{0.9\columnwidth}
        \centering
        \includegraphics[width=\linewidth]{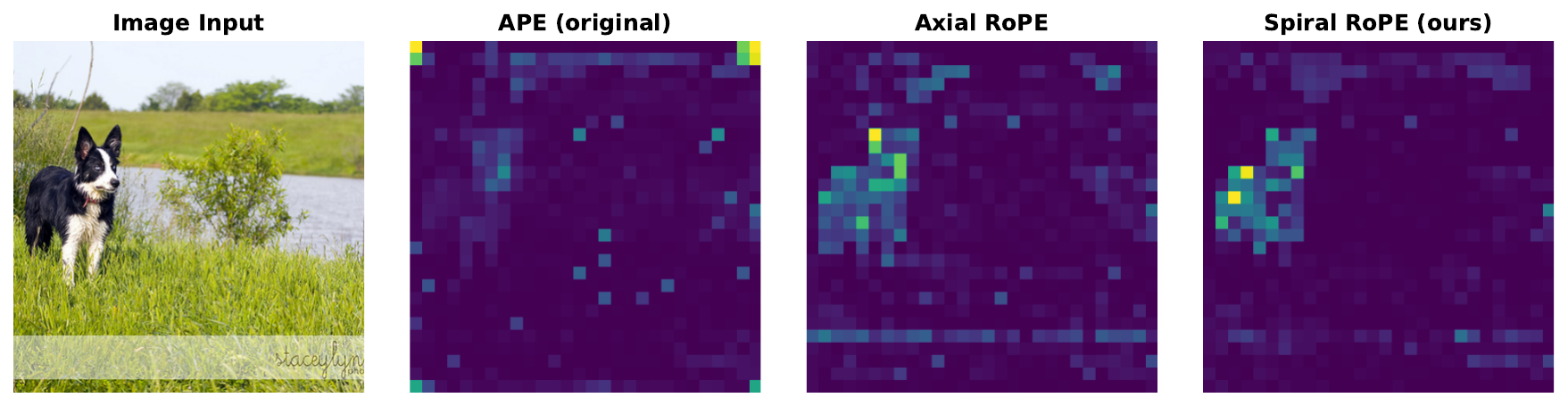}
    \end{subfigure}

    \caption{\textbf{\textit{Class-token attention visualization.}} Each row shows the input image followed by attention maps from APE, Axial 2D RoPE, and Spiral RoPE. Attention is averaged over all heads in the last layer. Spiral RoPE produces more object-centric attention with cleaner background suppression compared to both counterparts.}
    \label{fig:cls_attn}
\end{figure}

\paragraph{Class-Token Attention.} We first visualize the class-token attention by averaging the attention weights over all heads in the last transformer layer. \cref{fig:cls_attn} shows representative examples from the ImageNet validation set. A clear progression is visible across the three methods. The APE baseline produces diffuse attention patterns, with substantial activation spread across both foreground objects and background regions. Axial RoPE improves over APE by allocating stronger attention to semantically relevant regions; however, its attention maps still exhibit noticeable residual activations in surrounding background areas and irrelevant objects. 
In contrast, Spiral RoPE yields the most object-centric and spatially coherent attention patterns, characterized by concentrated responses on the foreground objects and effective suppression of background activations. 
This advantage is particularly pronounced in multi-instance scenes, where Spiral RoPE sharply attends to multiple relevant objects, while Axial RoPE still exhibits axis-aligned artifacts and attention leakage into nearby regions.

\paragraph{Query-Token Attention.} To further examine how positional encoding influences local spatial relationships, we visualize attention patterns originating from individual patch tokens. 
\cref{fig:query_attn} shows representative examples in which the query token (marked by a red box) is located on a foreground object. 
Across all examples, APE and Axial RoPE produce attention maps that are broadly distributed across the image, with substantial activation on background regions and unrelated objects. 
In contrast, Spiral RoPE consistently produces markedly sharper and more spatially coherent attention patterns, with strong responses concentrated around the query location and semantically related regions. 
This suggests that Spiral RoPE facilitates more precise spatial alignment between query and key tokens, enabling attention to better respect local object boundaries and spatial proximity. 
We attribute this behavior to the multi-directional nature of Spiral RoPE, which encodes spatial relationships beyond axis-aligned directions and thus supports more flexible two-dimensional spatial interactions.

\subsection{Ablation Studies}

We conduct ablation studies on the ImageNet-1k classification task using DeiT-Base at $224 \times 224$ resolution to analyze the effect of key hyperparameters in Spiral RoPE. All ablation experiments use the same training recipe as described in Section~\ref{sec:experiments:classification}, with evaluation using the final checkpoint with EMA weights and full crop.

\paragraph{Number of Directions $K$.} We first ablate the number of directions $K$ in Spiral RoPE with the frequency scaling factor fixed at 1.5. As shown in \cref{tab:ablation_k}, Spiral RoPE consistently outperforms the axial RoPE baseline (83.31\%) across all values of $K$. Increasing $K$ from 4 to 16 gradually improves accuracy, with $K=16$ achieving the best performance (83.39\%). However, further increasing $K$ to 32 slightly degrades performance (83.32\%), suggesting that excessive directional partitioning may reduce the embedding capacity per direction without providing additional benefit. Based on these results, we select $K=16$ as the default configuration for our classification experiments.

\paragraph{Frequency Scaling Factor.} We also ablate the frequency scaling factor that multiplies all base frequencies $\theta_t$ in Spiral RoPE, with $K$ fixed at 16. As shown in \cref{tab:ablation_freq}, the scaling factor has a notable impact on performance. A scaling factor of 1.0 (no scaling) achieves 83.33\%, while increasing to 1.5 yields the best result (83.39\%). However, larger scaling factors (1.75, 2.0) lead to decreased performance, likely because excessively high frequencies cause the rotations to vary too rapidly across positions, making it difficult for the model to learn stable positional relationships. We select 1.5 as the default scaling factor for our experiments.

\begin{table}[t]
    \centering
    \begin{minipage}[t]{0.48\columnwidth}
        \centering
        \caption{Ablation on the number of directions $K$.}
        \label{tab:ablation_k}
        \vspace{0.5em}
        \begin{small}
            \begin{sc}
                \begin{tabular}{lc}
                    \toprule
                    $K$ & Acc (\%) \\
                    \midrule
                    2 (Axial RoPE) & 83.31 \\
                    4 & 83.31 \\
                    8 & 83.34 \\
                    16 & \textbf{83.39} \\
                    32 & 83.32 \\
                    \bottomrule
                \end{tabular}
            \end{sc}
        \end{small}
    \end{minipage}
    \hfill
    \begin{minipage}[t]{0.48\columnwidth}
        \centering
        \caption{Ablation on the frequency scaling factor.}
        \label{tab:ablation_freq}
        \vspace{0.5em}
        \begin{small}
            \begin{sc}
                \begin{tabular}{lc}
                    \toprule
                    Scale & Acc (\%) \\
                    \midrule
                    1.00 & 83.33 \\
                    1.25 & 83.13 \\
                    1.50 & \textbf{83.39} \\
                    1.75 & 83.34 \\
                    2.00 & 83.23 \\
                    \bottomrule
                \end{tabular}
            \end{sc}
        \end{small}
    \end{minipage}
    \vskip -0.1in
\end{table}


\section{Conclusion}
\label{sec:conclusion}

We have presented Spiral RoPE, a simple yet effective extension of rotary position embedding for vision transformers that addresses the directional constraint of conventional Axial 2D RoPE by distributing positional encoding across $K$ uniformly spaced directions. Extensive experiments across image classification, semantic segmentation, and image generation demonstrate consistent performance improvements, with particularly strong resolution extrapolation capability. These results suggest that directional diversity in positional encoding plays a critical role in modeling 2D spatial relationships, and that modest geometric extensions to rotary embeddings can substantially enhance spatial representation in vision transformers.






\bibliography{example_paper}

@misc{baiQwenTechnicalReport2023,
  title = {Qwen {{Technical Report}}},
  year = 2023,
  month = sep,
  number = {arXiv:2309.16609},
  eprint = {2309.16609},
  publisher = {arXiv},
  doi = {10.48550/arXiv.2309.16609},
  url = {http://arxiv.org/abs/2309.16609},
  urldate = {2026-01-16},
  archiveprefix = {arXiv},
  author = {Bai, Jinze and Bai, Shuai and Chu, Yunfei and Cui, Zeyu and Dang, Kai and Deng, Xiaodong and Fan, Yang and Ge, Wenbin and Han, Yu and Huang, Fei and Hui, Binyuan and Ji, Luo and Li, Mei and Lin, Junyang and Lin, Runji and Liu, Dayiheng and Liu, Gao and Lu, Chengqiang and Lu, Keming and others}
}

@misc{chenExtendingContextWindow2023,
  title = {Extending {{Context Window}} of {{Large Language Models}} via {{Positional Interpolation}}},
  year = 2023,
  month = jun,
  number = {arXiv:2306.15595},
  eprint = {2306.15595},
  publisher = {arXiv},
  doi = {10.48550/arXiv.2306.15595},
  url = {http://arxiv.org/abs/2306.15595},
  urldate = {2025-12-30},
  archiveprefix = {arXiv},
  author = {Chen, Shouyuan and Wong, Sherman and Chen, Liangjian and Tian, Yuandong}
}

@inproceedings{dengImageNetLargescaleHierarchical2009,
  title = {{{ImageNet}}: {{A}} Large-Scale Hierarchical Image Database},
  shorttitle = {{{ImageNet}}},
  booktitle = {2009 {{IEEE Conference}} on {{Computer Vision}} and {{Pattern Recognition}}},
  year = 2009,
  month = jun,
  pages = {248--255},
  issn = {1063-6919},
  doi = {10.1109/CVPR.2009.5206848},
  url = {https://ieeexplore.ieee.org/document/5206848},
  urldate = {2025-12-29},
  author = {Deng, Jia and Dong, Wei and Socher, Richard and Li, Li-Jia and Li, Kai and {Fei-Fei}, Li}
}

@misc{devlinBERTPretrainingDeep2019,
  title = {{{BERT}}: {{Pre-training}} of {{Deep Bidirectional Transformers}} for {{Language Understanding}}},
  shorttitle = {{{BERT}}},
  year = 2019,
  month = may,
  number = {arXiv:1810.04805},
  eprint = {1810.04805},
  publisher = {arXiv},
  doi = {10.48550/arXiv.1810.04805},
  url = {http://arxiv.org/abs/1810.04805},
  urldate = {2025-12-30},
  archiveprefix = {arXiv},
  author = {Devlin, Jacob and Chang, Ming-Wei and Lee, Kenton and Toutanova, Kristina}
}

@misc{dhariwalDiffusionModelsBeat2021,
  title = {Diffusion {{Models Beat GANs}} on {{Image Synthesis}}},
  year = 2021,
  month = jun,
  number = {arXiv:2105.05233},
  eprint = {2105.05233},
  publisher = {arXiv},
  doi = {10.48550/arXiv.2105.05233},
  url = {http://arxiv.org/abs/2105.05233},
  urldate = {2026-01-05},
  archiveprefix = {arXiv},
  author = {Dhariwal, Prafulla and Nichol, Alex}
}

@misc{dosovitskiyImageWorth16x162021,
  title = {An {{Image}} Is {{Worth}} 16x16 {{Words}}: {{Transformers}} for {{Image Recognition}} at {{Scale}}},
  shorttitle = {An {{Image}} Is {{Worth}} 16x16 {{Words}}},
  year = 2021,
  month = jun,
  number = {arXiv:2010.11929},
  eprint = {2010.11929},
  publisher = {arXiv},
  doi = {10.48550/arXiv.2010.11929},
  url = {http://arxiv.org/abs/2010.11929},
  urldate = {2025-12-30},
  archiveprefix = {arXiv},
  author = {Dosovitskiy, Alexey and Beyer, Lucas and Kolesnikov, Alexander and Weissenborn, Dirk and Zhai, Xiaohua and Unterthiner, Thomas and Dehghani, Mostafa and Minderer, Matthias and Heigold, Georg and Gelly, Sylvain and Uszkoreit, Jakob and Houlsby, Neil}
}

@article{fangEVA02VisualRepresentation2024,
  title = {{{EVA-02}}: {{A Visual Representation}} for {{Neon Genesis}}},
  shorttitle = {{{EVA-02}}},
  year = 2024,
  month = sep,
  journal = {Image and Vision Computing},
  volume = {149},
  eprint = {2303.11331},
  pages = {105171},
  issn = {02628856},
  doi = {10.1016/j.imavis.2024.105171},
  url = {http://arxiv.org/abs/2303.11331},
  urldate = {2026-01-04},
  archiveprefix = {arXiv},
  author = {Fang, Yuxin and Sun, Quan and Wang, Xinggang and Huang, Tiejun and Wang, Xinlong and Cao, Yue}
}

@misc{grattafioriLlama3Herd2024,
  title = {The {{Llama}} 3 {{Herd}} of {{Models}}},
  year = 2024,
  month = nov,
  number = {arXiv:2407.21783},
  eprint = {2407.21783},
  publisher = {arXiv},
  doi = {10.48550/arXiv.2407.21783},
  url = {http://arxiv.org/abs/2407.21783},
  urldate = {2026-01-16},
  archiveprefix = {arXiv},
  author = {Grattafiori, Aaron and Dubey, Abhimanyu and Jauhri, Abhinav and Pandey, Abhinav and Kadian, Abhishek and {Al-Dahle}, Ahmad and Letman, Aiesha and Mathur, Akhil and Schelten, Alan and Vaughan, Alex and Yang, Amy and Fan, Angela and Goyal, Anirudh and Hartshorn, Anthony and Yang, Aobo and Mitra, Archi and Sravankumar, Archie and Korenev, Artem and Hinsvark, Arthur and others}
}

@inproceedings{heoRotaryPositionEmbedding2024,
  title = {Rotary {{Position Embedding}} for~{{Vision Transformer}}},
  booktitle = {Computer {{Vision}} – {{ECCV}} 2024: 18th {{European Conference}}, {{Milan}}, {{Italy}}, {{September}} 29–{{October}} 4, 2024, {{Proceedings}}, {{Part X}}},
  year = 2024,
  month = nov,
  pages = {289--305},
  publisher = {Springer-Verlag},
  address = {Berlin, Heidelberg},
  doi = {10.1007/978-3-031-72684-2_17},
  url = {https://doi.org/10.1007/978-3-031-72684-2_17},
  urldate = {2025-12-26},
  isbn = {978-3-031-72683-5},
  author = {Heo, Byeongho and Park, Song and Han, Dongyoon and Yun, Sangdoo}
}

@misc{heuselGANsTrainedTwo2018,
  title = {{{GANs Trained}} by a {{Two Time-Scale Update Rule Converge}} to a {{Local Nash Equilibrium}}},
  year = 2018,
  month = jan,
  number = {arXiv:1706.08500},
  eprint = {1706.08500},
  publisher = {arXiv},
  doi = {10.48550/arXiv.1706.08500},
  url = {http://arxiv.org/abs/1706.08500},
  urldate = {2025-12-30},
  archiveprefix = {arXiv},
  author = {Heusel, Martin and Ramsauer, Hubert and Unterthiner, Thomas and Nessler, Bernhard and Hochreiter, Sepp}
}

@misc{liuSwinTransformerHierarchical2021,
  title = {Swin {{Transformer}}: {{Hierarchical Vision Transformer}} Using {{Shifted Windows}}},
  shorttitle = {Swin {{Transformer}}},
  year = 2021,
  month = aug,
  number = {arXiv:2103.14030},
  eprint = {2103.14030},
  publisher = {arXiv},
  doi = {10.48550/arXiv.2103.14030},
  url = {http://arxiv.org/abs/2103.14030},
  urldate = {2025-12-30},
  archiveprefix = {arXiv},
  author = {Liu, Ze and Lin, Yutong and Cao, Yue and Hu, Han and Wei, Yixuan and Zhang, Zheng and Lin, Stephen and Guo, Baining}
}

@misc{luFiTFlexibleVision2024,
  title = {{{FiT}}: {{Flexible Vision Transformer}} for {{Diffusion Model}}},
  shorttitle = {{{FiT}}},
  year = 2024,
  month = oct,
  number = {arXiv:2402.12376},
  eprint = {2402.12376},
  publisher = {arXiv},
  doi = {10.48550/arXiv.2402.12376},
  url = {http://arxiv.org/abs/2402.12376},
  urldate = {2026-01-04},
  archiveprefix = {arXiv},
  author = {Lu, Zeyu and Wang, Zidong and Huang, Di and Wu, Chengyue and Liu, Xihui and Ouyang, Wanli and Bai, Lei}
}

@misc{luUnifiedIO2Scaling2023,
  title = {Unified-{{IO}} 2: {{Scaling Autoregressive Multimodal Models}} with {{Vision}}, {{Language}}, {{Audio}}, and {{Action}}},
  shorttitle = {Unified-{{IO}} 2},
  year = 2023,
  month = dec,
  number = {arXiv:2312.17172},
  eprint = {2312.17172},
  publisher = {arXiv},
  doi = {10.48550/arXiv.2312.17172},
  url = {http://arxiv.org/abs/2312.17172},
  urldate = {2026-01-04},
  archiveprefix = {arXiv},
  author = {Lu, Jiasen and Clark, Christopher and Lee, Sangho and Zhang, Zichen and Khosla, Savya and Marten, Ryan and Hoiem, Derek and Kembhavi, Aniruddha}
}

@misc{maSiTExploringFlow2024,
  title = {{{SiT}}: {{Exploring Flow}} and {{Diffusion-based Generative Models}} with {{Scalable Interpolant Transformers}}},
  shorttitle = {{{SiT}}},
  year = 2024,
  month = sep,
  number = {arXiv:2401.08740},
  eprint = {2401.08740},
  publisher = {arXiv},
  doi = {10.48550/arXiv.2401.08740},
  url = {http://arxiv.org/abs/2401.08740},
  urldate = {2026-01-20},
  archiveprefix = {arXiv},
  author = {Ma, Nanye and Goldstein, Mark and Albergo, Michael S. and Boffi, Nicholas M. and {Vanden-Eijnden}, Eric and Xie, Saining}
}

@misc{peeblesScalableDiffusionModels2023,
  title = {Scalable {{Diffusion Models}} with {{Transformers}}},
  year = 2023,
  month = mar,
  number = {arXiv:2212.09748},
  eprint = {2212.09748},
  publisher = {arXiv},
  doi = {10.48550/arXiv.2212.09748},
  url = {http://arxiv.org/abs/2212.09748},
  urldate = {2025-12-26},
  archiveprefix = {arXiv},
  author = {Peebles, William and Xie, Saining}
}

@misc{pengYaRNEfficientContext2023,
  title = {{{YaRN}}: {{Efficient Context Window Extension}} of {{Large Language Models}}},
  shorttitle = {{{YaRN}}},
  year = 2023,
  month = nov,
  number = {arXiv:2309.00071},
  eprint = {2309.00071},
  publisher = {arXiv},
  doi = {10.48550/arXiv.2309.00071},
  url = {http://arxiv.org/abs/2309.00071},
  urldate = {2025-12-30},
  archiveprefix = {arXiv},
  author = {Peng, Bowen and Quesnelle, Jeffrey and Fan, Honglu and Shippole, Enrico}
}

@misc{pressTrainShortTest2022,
  title = {Train {{Short}}, {{Test Long}}: {{Attention}} with {{Linear Biases Enables Input Length Extrapolation}}},
  shorttitle = {Train {{Short}}, {{Test Long}}},
  year = 2022,
  month = apr,
  number = {arXiv:2108.12409},
  eprint = {2108.12409},
  publisher = {arXiv},
  doi = {10.48550/arXiv.2108.12409},
  url = {http://arxiv.org/abs/2108.12409},
  urldate = {2025-12-30},
  archiveprefix = {arXiv},
  author = {Press, Ofir and Smith, Noah A. and Lewis, Mike}
}

@misc{raffelExploringLimitsTransfer2023,
  title = {Exploring the {{Limits}} of {{Transfer Learning}} with a {{Unified Text-to-Text Transformer}}},
  year = 2023,
  month = sep,
  number = {arXiv:1910.10683},
  eprint = {1910.10683},
  publisher = {arXiv},
  doi = {10.48550/arXiv.1910.10683},
  url = {http://arxiv.org/abs/1910.10683},
  urldate = {2025-12-30},
  archiveprefix = {arXiv},
  author = {Raffel, Colin and Shazeer, Noam and Roberts, Adam and Lee, Katherine and Narang, Sharan and Matena, Michael and Zhou, Yanqi and Li, Wei and Liu, Peter J.}
}

@misc{raviSAM2Segment2024,
  title = {{{SAM}} 2: {{Segment Anything}} in {{Images}} and {{Videos}}},
  shorttitle = {{{SAM}} 2},
  year = 2024,
  month = oct,
  number = {arXiv:2408.00714},
  eprint = {2408.00714},
  publisher = {arXiv},
  doi = {10.48550/arXiv.2408.00714},
  url = {http://arxiv.org/abs/2408.00714},
  urldate = {2025-12-30},
  archiveprefix = {arXiv},
  author = {Ravi, Nikhila and Gabeur, Valentin and Hu, Yuan-Ting and Hu, Ronghang and Ryali, Chaitanya and Ma, Tengyu and Khedr, Haitham and Rädle, Roman and Rolland, Chloe and Gustafson, Laura and Mintun, Eric and Pan, Junting and Alwala, Kalyan Vasudev and Carion, Nicolas and Wu, Chao-Yuan and Girshick, Ross and Dollár, Piotr and Feichtenhofer, Christoph}
}

@misc{shawSelfAttentionRelativePosition2018,
  title = {Self-{{Attention}} with {{Relative Position Representations}}},
  year = 2018,
  month = apr,
  number = {arXiv:1803.02155},
  eprint = {1803.02155},
  publisher = {arXiv},
  doi = {10.48550/arXiv.1803.02155},
  url = {http://arxiv.org/abs/1803.02155},
  urldate = {2025-12-30},
  archiveprefix = {arXiv},
  author = {Shaw, Peter and Uszkoreit, Jakob and Vaswani, Ashish}
}

@misc{suRoFormerEnhancedTransformer2023,
  title = {{{RoFormer}}: {{Enhanced Transformer}} with {{Rotary Position Embedding}}},
  shorttitle = {{{RoFormer}}},
  year = 2023,
  month = nov,
  number = {arXiv:2104.09864},
  eprint = {2104.09864},
  publisher = {arXiv},
  doi = {10.48550/arXiv.2104.09864},
  url = {http://arxiv.org/abs/2104.09864},
  urldate = {2025-12-26},
  archiveprefix = {arXiv},
  author = {Su, Jianlin and Lu, Yu and Pan, Shengfeng and Murtadha, Ahmed and Wen, Bo and Liu, Yunfeng}
}

@misc{tianUDiTsDownsampleTokens2024,
  title = {U-{{DiTs}}: {{Downsample Tokens}} in {{U-Shaped Diffusion Transformers}}},
  shorttitle = {U-{{DiTs}}},
  year = 2024,
  month = oct,
  number = {arXiv:2405.02730},
  eprint = {2405.02730},
  publisher = {arXiv},
  doi = {10.48550/arXiv.2405.02730},
  url = {http://arxiv.org/abs/2405.02730},
  urldate = {2026-01-14},
  archiveprefix = {arXiv},
  author = {Tian, Yuchuan and Tu, Zhijun and Chen, Hanting and Hu, Jie and Xu, Chao and Wang, Yunhe}
}

@misc{touvronLlama2Open2023,
  title = {Llama 2: {{Open Foundation}} and {{Fine-Tuned Chat Models}}},
  shorttitle = {Llama 2},
  year = 2023,
  month = jul,
  number = {arXiv:2307.09288},
  eprint = {2307.09288},
  publisher = {arXiv},
  doi = {10.48550/arXiv.2307.09288},
  url = {http://arxiv.org/abs/2307.09288},
  urldate = {2026-01-16},
  archiveprefix = {arXiv},
  author = {Touvron, Hugo and Martin, Louis and Stone, Kevin and Albert, Peter and Almahairi, Amjad and Babaei, Yasmine and Bashlykov, Nikolay and Batra, Soumya and Bhargava, Prajjwal and Bhosale, Shruti and Bikel, Dan and Blecher, Lukas and Ferrer, Cristian Canton and Chen, Moya and Cucurull, Guillem and Esiobu, David and Fernandes, Jude and Fu, Jeremy and Fu, Wenyin and others}
}

@misc{touvronLLaMAOpenEfficient2023,
  title = {{{LLaMA}}: {{Open}} and {{Efficient Foundation Language Models}}},
  shorttitle = {{{LLaMA}}},
  year = 2023,
  month = feb,
  number = {arXiv:2302.13971},
  eprint = {2302.13971},
  publisher = {arXiv},
  doi = {10.48550/arXiv.2302.13971},
  url = {http://arxiv.org/abs/2302.13971},
  urldate = {2025-12-30},
  archiveprefix = {arXiv},
  author = {Touvron, Hugo and Lavril, Thibaut and Izacard, Gautier and Martinet, Xavier and Lachaux, Marie-Anne and Lacroix, Timothée and Rozière, Baptiste and Goyal, Naman and Hambro, Eric and Azhar, Faisal and Rodriguez, Aurelien and Joulin, Armand and Grave, Edouard and Lample, Guillaume}
}

@misc{touvronTrainingDataefficientImage2021,
  title = {Training Data-Efficient Image Transformers \& Distillation through Attention},
  year = 2021,
  month = jan,
  number = {arXiv:2012.12877},
  eprint = {2012.12877},
  publisher = {arXiv},
  doi = {10.48550/arXiv.2012.12877},
  url = {http://arxiv.org/abs/2012.12877},
  urldate = {2025-12-26},
  archiveprefix = {arXiv},
  author = {Touvron, Hugo and Cord, Matthieu and Douze, Matthijs and Massa, Francisco and Sablayrolles, Alexandre and Jégou, Hervé}
}

@misc{vaswaniAttentionAllYou2023,
  title = {Attention {{Is All You Need}}},
  year = 2023,
  month = aug,
  number = {arXiv:1706.03762},
  eprint = {1706.03762},
  publisher = {arXiv},
  doi = {10.48550/arXiv.1706.03762},
  url = {http://arxiv.org/abs/1706.03762},
  urldate = {2025-12-30},
  archiveprefix = {arXiv},
  author = {Vaswani, Ashish and Shazeer, Noam and Parmar, Niki and Uszkoreit, Jakob and Jones, Llion and Gomez, Aidan N. and Kaiser, Lukasz and Polosukhin, Illia}
}

@misc{wangCircleRoPEConelikeDecoupled2025,
  title = {Circle-{{RoPE}}: {{Cone-like Decoupled Rotary Positional Embedding}} for {{Large Vision-Language Models}}},
  shorttitle = {Circle-{{RoPE}}},
  year = 2025,
  month = oct,
  number = {arXiv:2505.16416},
  eprint = {2505.16416},
  publisher = {arXiv},
  doi = {10.48550/arXiv.2505.16416},
  url = {http://arxiv.org/abs/2505.16416},
  urldate = {2025-12-30},
  archiveprefix = {arXiv},
  author = {Wang, Chengcheng and Guo, Jianyuan and Li, Hongguang and Tian, Yuchuan and Nie, Ying and Xu, Chang and Han, Kai}
}

@misc{wangQwen2VLEnhancingVisionLanguage2024,
  title = {Qwen2-{{VL}}: {{Enhancing Vision-Language Model}}'s {{Perception}} of the {{World}} at {{Any Resolution}}},
  shorttitle = {Qwen2-{{VL}}},
  year = 2024,
  month = sep,
  number = {arXiv:2409.12191},
  eprint = {2409.12191},
  publisher = {arXiv},
  doi = {10.48550/arXiv.2409.12191},
  url = {http://arxiv.org/abs/2409.12191},
  urldate = {2025-12-30},
  archiveprefix = {arXiv},
  author = {Wang, Peng and Bai, Shuai and Tan, Sinan and Wang, Shijie and Fan, Zhihao and Bai, Jinze and Chen, Keqin and Liu, Xuejing and Wang, Jialin and Ge, Wenbin and Fan, Yang and Dang, Kai and Du, Mengfei and Ren, Xuancheng and Men, Rui and Liu, Dayiheng and Zhou, Chang and Zhou, Jingren and Lin, Junyang}
}

@misc{wuQwenImageTechnicalReport2025,
  title = {Qwen-{{Image Technical Report}}},
  year = 2025,
  month = aug,
  number = {arXiv:2508.02324},
  eprint = {2508.02324},
  publisher = {arXiv},
  doi = {10.48550/arXiv.2508.02324},
  url = {http://arxiv.org/abs/2508.02324},
  urldate = {2025-12-30},
  archiveprefix = {arXiv},
  author = {Wu, Chenfei and Li, Jiahao and Zhou, Jingren and Lin, Junyang and Gao, Kaiyuan and Yan, Kun and Yin, Sheng-ming and Bai, Shuai and Xu, Xiao and Chen, Yilei and Chen, Yuxiang and Tang, Zecheng and Zhang, Zekai and Wang, Zhengyi and Yang, An and Yu, Bowen and Cheng, Chen and Liu, Dayiheng and Li, Deqing and others}
}

@misc{xiaoUnifiedPerceptualParsing2018,
  title = {Unified {{Perceptual Parsing}} for {{Scene Understanding}}},
  year = 2018,
  month = jul,
  number = {arXiv:1807.10221},
  eprint = {1807.10221},
  publisher = {arXiv},
  doi = {10.48550/arXiv.1807.10221},
  url = {http://arxiv.org/abs/1807.10221},
  urldate = {2025-12-30},
  archiveprefix = {arXiv},
  author = {Xiao, Tete and Liu, Yingcheng and Zhou, Bolei and Jiang, Yuning and Sun, Jian}
}

@misc{yangQwen25TechnicalReport2025,
  title = {Qwen2.5 {{Technical Report}}},
  year = 2025,
  month = jan,
  number = {arXiv:2412.15115},
  eprint = {2412.15115},
  publisher = {arXiv},
  doi = {10.48550/arXiv.2412.15115},
  url = {http://arxiv.org/abs/2412.15115},
  urldate = {2026-01-16},
  archiveprefix = {arXiv},
  author = {Yang, An and Zhang, Beichen and Hui, Binyuan and Zheng, Bo and Yu, Bowen and Li, Chengyuan and Liu, Dayiheng and Huang, Fei and Wei, Haoran and Lin, Huan and Yang, Jian and Tu, Jianhong and Zhang, Jianwei and Yang, Jianxin and Yang, Jiaxi and Zhou, Jingren and Lin, Junyang and Dang, Kai and Lu, Keming and others}
}

@misc{yangQwen2TechnicalReport2024,
  title = {Qwen2 {{Technical Report}}},
  year = 2024,
  month = sep,
  number = {arXiv:2407.10671},
  eprint = {2407.10671},
  publisher = {arXiv},
  doi = {10.48550/arXiv.2407.10671},
  url = {http://arxiv.org/abs/2407.10671},
  urldate = {2026-01-16},
  archiveprefix = {arXiv},
  author = {Yang, An and Yang, Baosong and Hui, Binyuan and Zheng, Bo and Yu, Bowen and Zhou, Chang and Li, Chengpeng and Li, Chengyuan and Liu, Dayiheng and Huang, Fei and Dong, Guanting and Wei, Haoran and Lin, Huan and Tang, Jialong and Wang, Jialin and Yang, Jian and Tu, Jianhong and Zhang, Jianwei and Ma, Jianxin and others}
}

@misc{yangQwen3TechnicalReport2025,
  title = {Qwen3 {{Technical Report}}},
  year = 2025,
  month = may,
  number = {arXiv:2505.09388},
  eprint = {2505.09388},
  publisher = {arXiv},
  doi = {10.48550/arXiv.2505.09388},
  url = {http://arxiv.org/abs/2505.09388},
  urldate = {2025-12-30},
  archiveprefix = {arXiv},
  author = {Yang, An and Li, Anfeng and Yang, Baosong and Zhang, Beichen and Hui, Binyuan and Zheng, Bo and Yu, Bowen and Gao, Chang and Huang, Chengen and Lv, Chenxu and Zheng, Chujie and Liu, Dayiheng and Zhou, Fan and Huang, Fei and Hu, Feng and Ge, Hao and Wei, Haoran and Lin, Huan and Tang, Jialong and others}
}

@inproceedings{zhouSceneParsingADE20K2017a,
  title = {Scene {{Parsing}} through {{ADE20K Dataset}}},
  booktitle = {2017 {{IEEE Conference}} on {{Computer Vision}} and {{Pattern Recognition}} ({{CVPR}})},
  year = 2017,
  month = jul,
  pages = {5122--5130},
  issn = {1063-6919},
  doi = {10.1109/CVPR.2017.544},
  url = {https://ieeexplore.ieee.org/document/8100027},
  urldate = {2025-12-30},
  author = {Zhou, Bolei and Zhao, Hang and Puig, Xavier and Fidler, Sanja and Barriuso, Adela and Torralba, Antonio}
}

@misc{zhouSemanticUnderstandingScenes2018a,
  title = {Semantic {{Understanding}} of {{Scenes}} through the {{ADE20K Dataset}}},
  year = 2018,
  month = oct,
  number = {arXiv:1608.05442},
  eprint = {1608.05442},
  publisher = {arXiv},
  doi = {10.48550/arXiv.1608.05442},
  url = {http://arxiv.org/abs/1608.05442},
  urldate = {2025-12-30},
  archiveprefix = {arXiv},
  author = {Zhou, Bolei and Zhao, Hang and Puig, Xavier and Xiao, Tete and Fidler, Sanja and Barriuso, Adela and Torralba, Antonio}
}
\bibliographystyle{icml2026}

\newpage

\appendix
\onecolumn

\section{Implementation Details}
\label{app:details}

\subsection{Image Classification}

\paragraph{Pre-training.} We train DeiT models on ImageNet-1K following an improved training recipe. For DeiT-Base and DeiT-Small, we train for 300 epochs with a batch size of 4096, learning rate of $2 \times 10^{-4}$, weight decay of 0.3, and drop path rate of 0.1. For DeiT-Large, we train for 200 epochs with the same hyperparameters except drop path rate of 0.2. We use AdamW optimizer with $\beta_1 = 0.95$ and cosine learning rate schedule with 20 warmup epochs. We employ the same data augmentation pipeline during training as the original DeiT implementation.

\paragraph{Fine-tuning.} For fine-tuning at $224 \times 224$, we train for 20 epochs with learning rate $10^{-5}$, weight decay 0.1, and the same drop path rate as pre-training. For higher resolutions ($384 \times 384$ and $448 \times 448$), we fine-tune from the $224 \times 224$ checkpoint for 20 epochs with the same hyperparameters. We use full crop (crop ratio 1.0) during evaluation.

\paragraph{Spiral RoPE Configuration.} For all classification experiments, we use $K=16$ directions with base frequency scaling factor 1.5 (i.e., $\theta_{\text{base}} = 15000$). Spiral RoPE is combined with the original absolute positional embedding (APE).

\begin{table}[h]
    \caption{Pre-training hyperparameters for ImageNet-1K classification.}
    \label{tab:pretrain_hyperparams}
    \begin{center}
        \begin{tabular}{lcc}
            \toprule
            Hyperparameter & DeiT-Base and DeiT-Small & DeiT-Large \\
            \midrule
            Epochs & 300 & 200 \\
            Batch size & 4096 & 4096 \\
            Optimizer & AdamW & AdamW \\
            Learning rate & $2 \times 10^{-4}$ & $2 \times 10^{-4}$ \\
            Weight decay & 0.3 & 0.3 \\
            $\beta_1$, $\beta_2$ & 0.95, 0.999 & 0.95, 0.999 \\
            LR schedule & Cosine & Cosine \\
            Warmup epochs & 20 & 20 \\
            Drop path & 0.1 & 0.2 \\
            Label smoothing & 0.1 & 0.1 \\
            \midrule
            \multicolumn{3}{c}{\textit{Data Augmentation}} \\
            \midrule
            RandAugment & (9, 0.5) & (9, 0.5) \\
            Mixup $\alpha$ & 0.8 & 0.8 \\
            CutMix $\alpha$ & 1.0 & 1.0 \\
            Random erasing & 0.25 & 0.25 \\
            \midrule
            \multicolumn{3}{c}{\textit{Spiral RoPE Parameters}} \\
            \midrule
            Number of directions ($K$) & 16 & 16 \\
            Base frequency scale & 1.5 & 1.5 \\
            \bottomrule
        \end{tabular}
    \end{center}
\end{table}

\subsection{Semantic Segmentation}
We use UperNet as the segmentation framework with DeiT backbones initialized from ImageNet pre-trained weights at $224 \times 224$ resolution. Models are trained on ADE20K at $512 \times 512$ resolution for 160K iterations with a total batch size of 16 (8 GPUs $\times$ 2 samples per GPU). We use AdamW optimizer with learning rate $6 \times 10^{-5}$, weight decay 0.01, and polynomial learning rate decay with 1.5K warmup iterations. Weight decay is disabled for position embeddings, class tokens, normalization layers, and layer scale parameters. For Spiral RoPE models, we use the same $K=16$ and base frequency scale 1.5 as in classification.

\subsection{Image Generation}

We follow the training recipe of the original DiT implementation \cite{peeblesScalableDiffusionModels2023}. Specifically, we train DiT models on ImageNet $256 \times 256$ for class-conditional image generation. All models are trained for 400K steps with a global batch size of 256. We use the AdamW optimizer with a constant learning rate of $10^{-4}$ and no weight decay. We employ the EMA of model weights with decay 0.9999 for evaluation. For sampling, we use DDPM with 250 steps and classifier-free guidance scale of 1.0.
FID is computed on 50K generated samples using the ADM evaluation toolkit \cite{dhariwalDiffusionModelsBeat2021}.

\paragraph{Spiral RoPE Configuration.} 
For DiT experiments, we use $K=8$ for S, B, and L models, and $K=6$ for XL models. The base frequency scale is set to 1.5 for all models.

\begin{table}[h]
    \caption{DiT training hyperparameters.}
    \label{tab:dit_hyperparams}
    \begin{center}
        \begin{tabular}{lc}
            \toprule
            Hyperparameter & Value \\
            \midrule
            Training steps & 400K \\
            Global batch size & 256 \\
            Optimizer & AdamW \\
            Learning rate & $1 \times 10^{-4}$ \\
            Weight decay & 0.0 \\
            EMA decay & 0.9999 \\
            \midrule
            \multicolumn{2}{c}{\textit{Sampling}} \\
            \midrule
            Sampling steps & 250 \\
            CFG scale & 1.0 \\
            \midrule
            \multicolumn{2}{c}{\textit{Spiral RoPE Parameters}} \\
            \midrule
            $K$ (S/B/L) & 8 \\
            $K$ (XL) & 6 \\
            Base frequency scale & 1.5 \\
            \bottomrule
        \end{tabular}
    \end{center}
\end{table}

\section{Additional Generated Samples}
\label{app:samples}

We provide additional uncurated samples generated by our DiT-XL/2 model with Spiral RoPE, trained for 7 million steps on ImageNet $256 \times 256$. For each class, we show 40 randomly generated samples. All samples are generated using classifier-free guidance with scale 4.0.

\begin{figure}[h]
    \centering
    \includegraphics[width=0.9\textwidth]{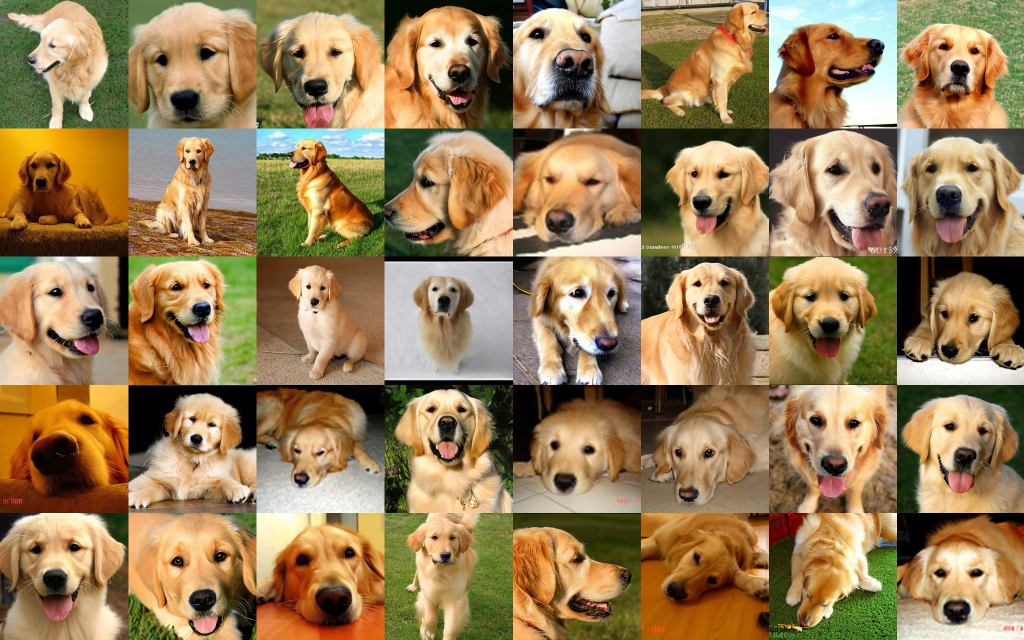}
    \caption{Class 207: Golden Retriever}
\end{figure}

\begin{figure}[h]
    \centering
    \includegraphics[width=0.9\textwidth]{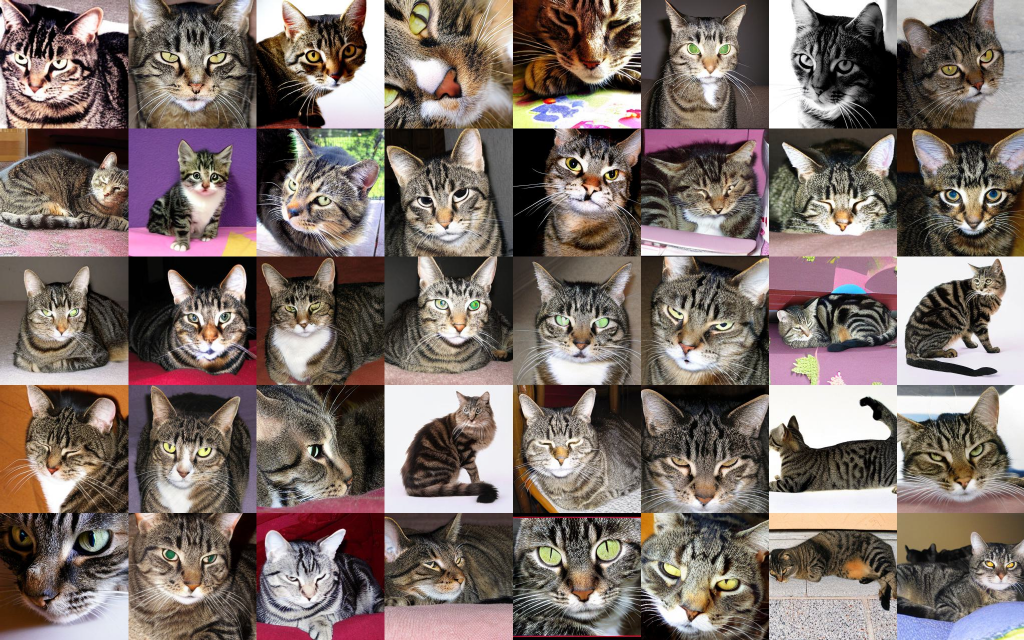}
    \caption{Class 281: Tabby Cat}
\end{figure}

\begin{figure}[h]
    \centering
    \includegraphics[width=0.9\textwidth]{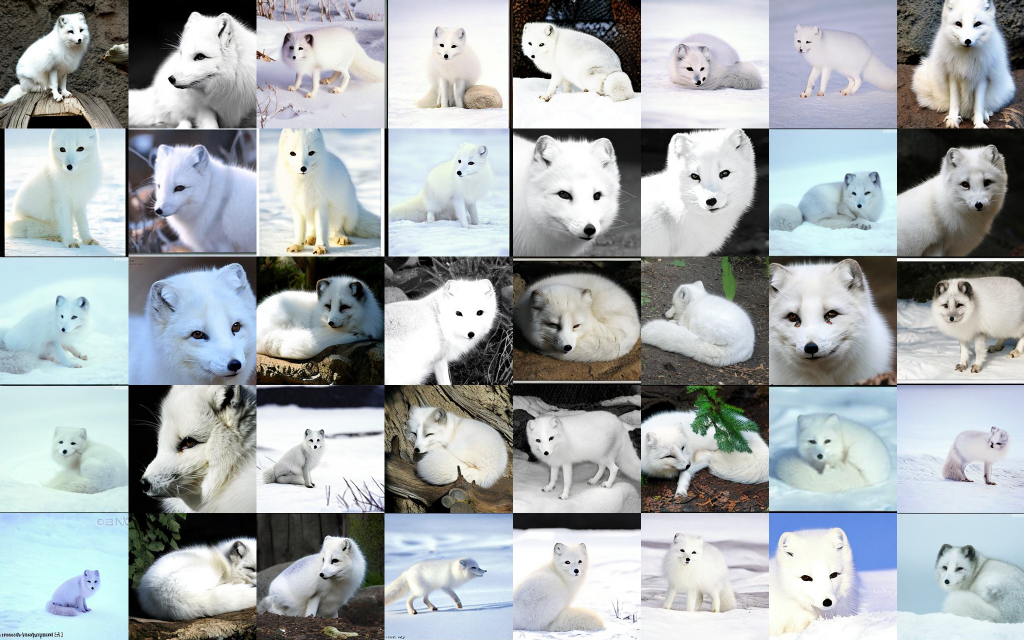}
    \caption{Class 279: Arctic Fox}
\end{figure}

\begin{figure}[h]
    \centering
    \includegraphics[width=0.9\textwidth]{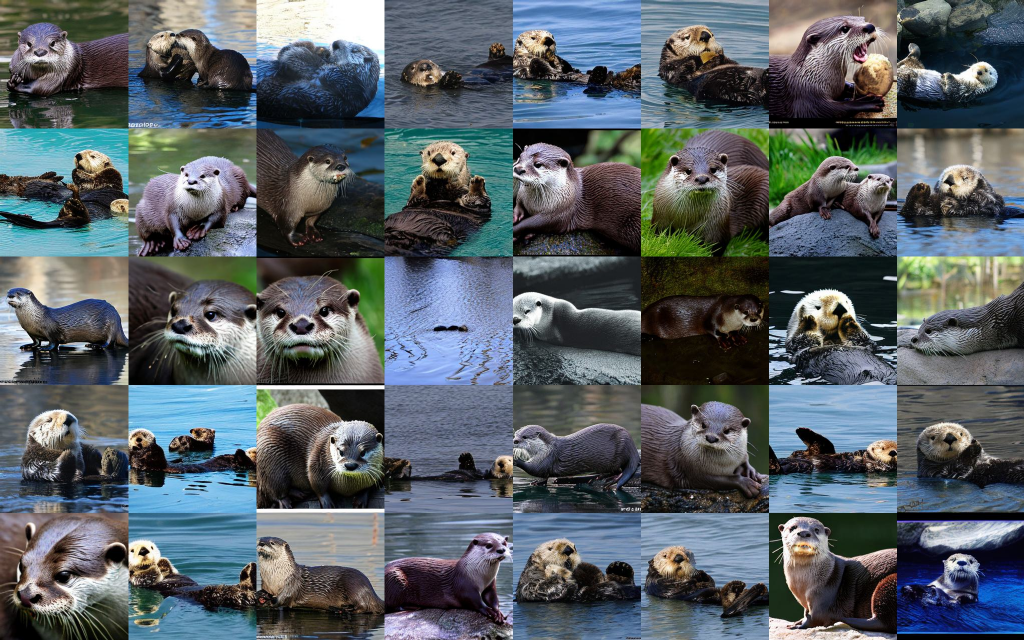}
    \caption{Class 360: Otter}
\end{figure}

\begin{figure}[h]
    \centering
    \includegraphics[width=0.9\textwidth]{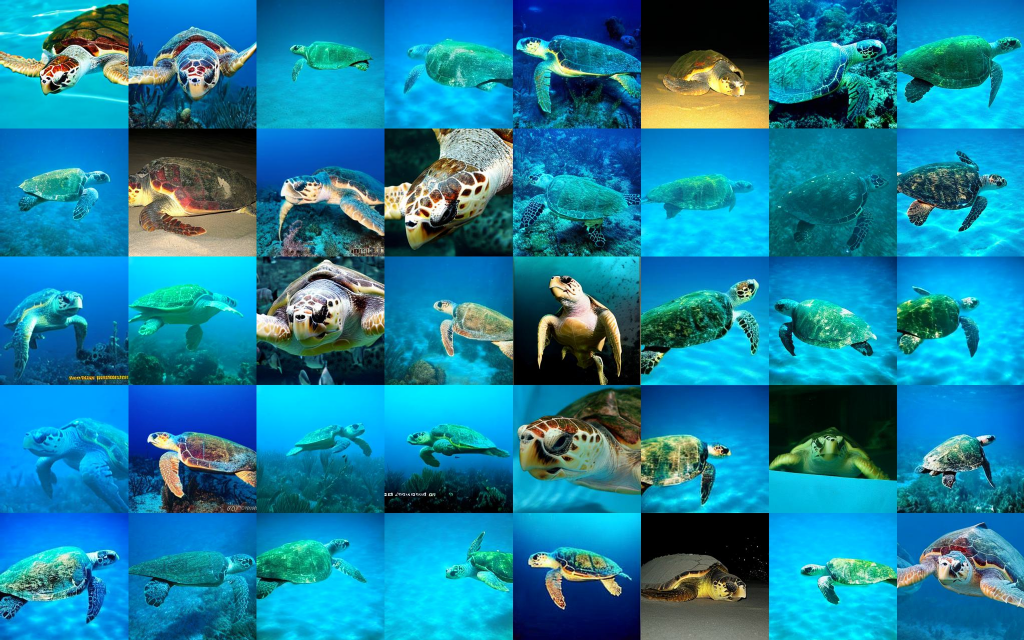}
    \caption{Class 33: Loggerhead Turtle}
\end{figure}

\begin{figure}[h]
    \centering
    \includegraphics[width=0.9\textwidth]{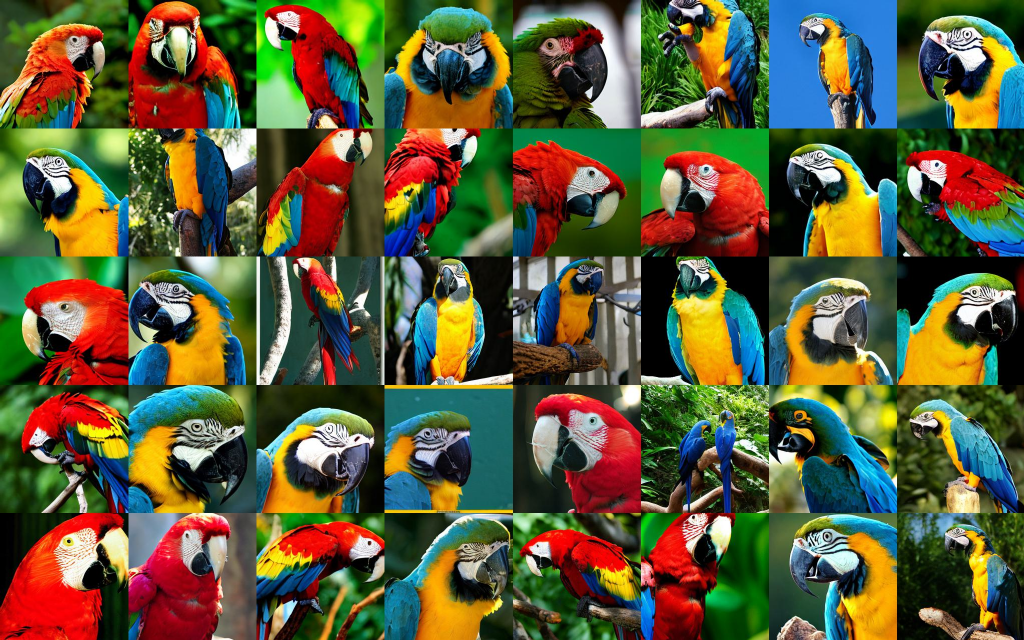}
    \caption{Class 88: Macaw}
\end{figure}

\begin{figure}[h]
    \centering
    \includegraphics[width=0.9\textwidth]{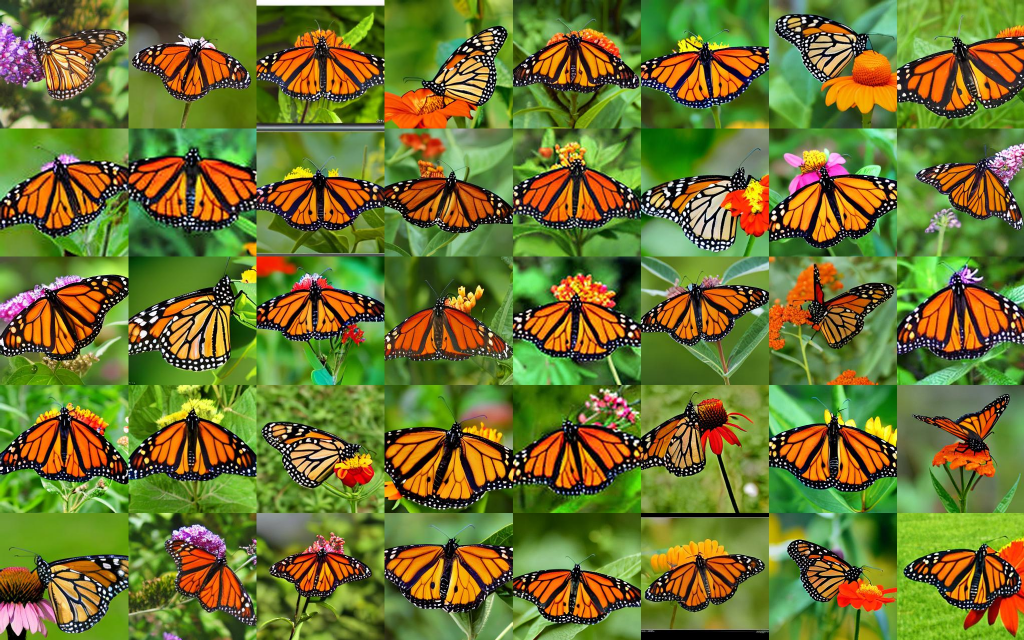}
    \caption{Class 323: Monarch Butterfly}
\end{figure}

\begin{figure}[h]
    \centering
    \includegraphics[width=0.9\textwidth]{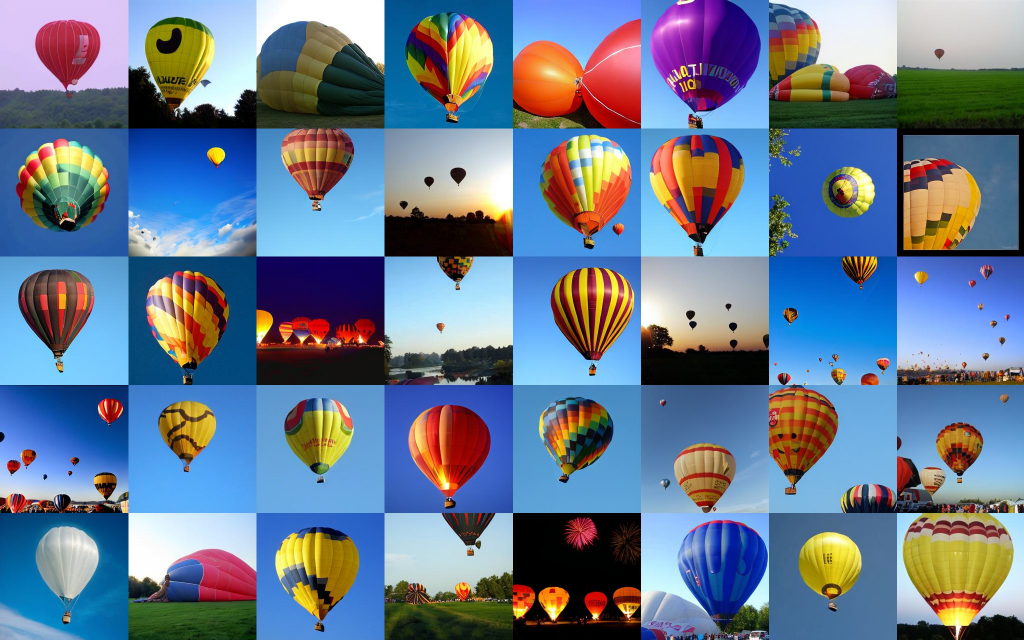}
    \caption{Class 417: Balloon}
\end{figure}

\begin{figure}[h]
    \centering
    \includegraphics[width=0.9\textwidth]{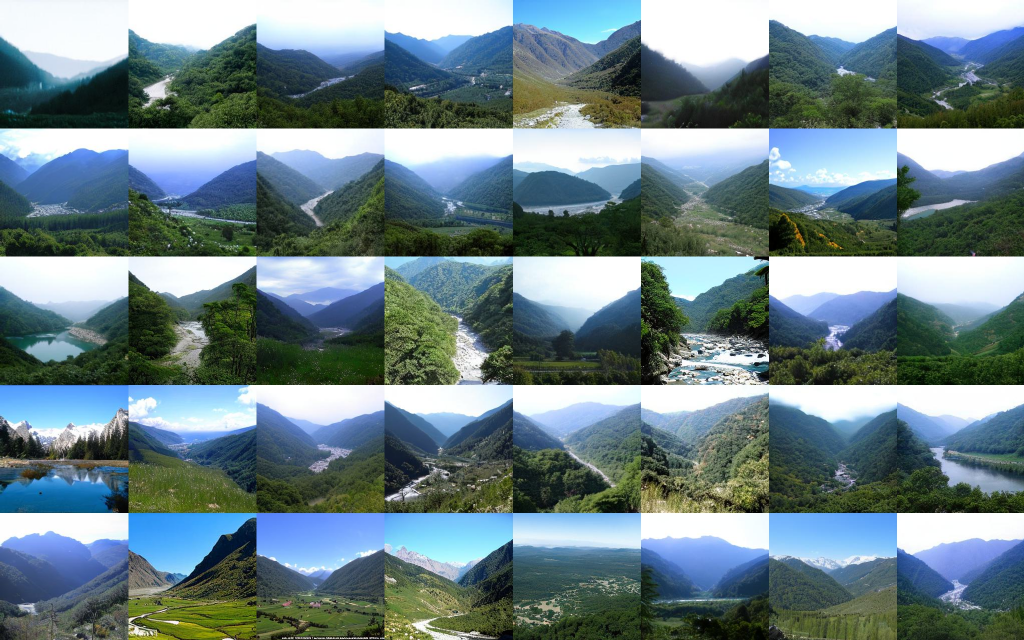}
    \caption{Class 979: Valley}
\end{figure}

\begin{figure}[h]
    \centering
    \includegraphics[width=0.9\textwidth]{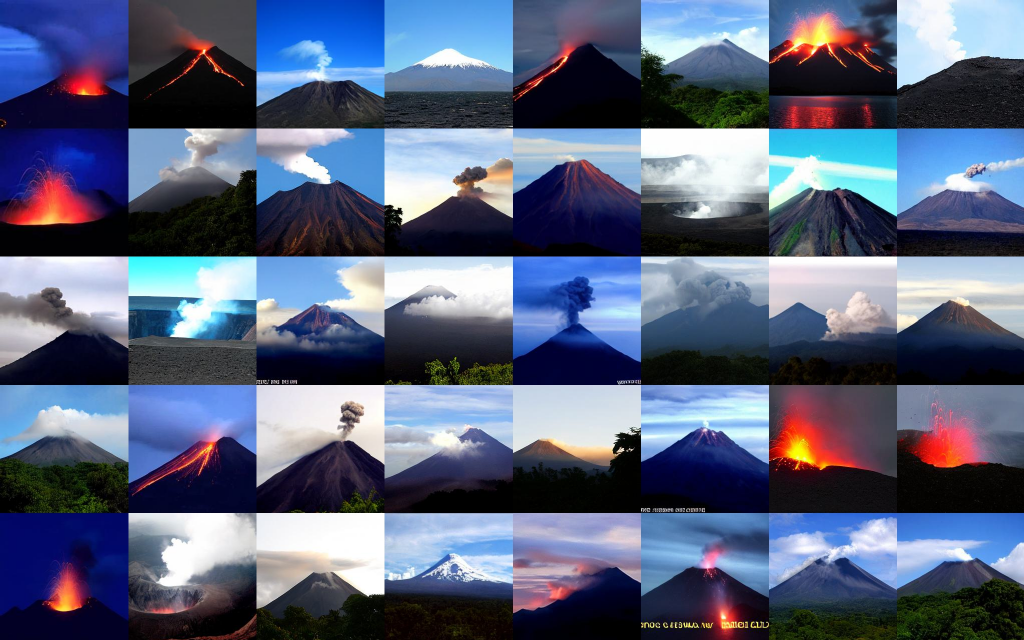}
    \caption{Class 980: Volcano}
\end{figure}

\end{document}